\def\year{2021}\relax
\documentclass[letterpaper]{article} 
\usepackage{aaai21}  
\usepackage{times}  
\usepackage{helvet} 
\usepackage{courier}  
\usepackage[hyphens]{url}  
\usepackage{graphicx} 
\usepackage{natbib}  
\usepackage{caption} 
\usepackage{booktabs}       
\usepackage{nicefrac}       
\usepackage{microtype}      
\usepackage{multirow}
\usepackage{colortbl}
\usepackage{color, colortbl}
\definecolor{Gray}{gray}{0.9}

\usepackage{tabulary}
\usepackage{booktabs}       
\usepackage[export]{adjustbox}

\usepackage{graphics,graphicx} 
\usepackage{subcaption} 

\usepackage{amsmath,amsfonts,bm}

\def\eqref#1{equation~\ref{#1}}

\def\1{\bm{1}}

\def\rvu{{\mathbf{i}}}

\def\rvu{{\mathbf{u}}}
\def\rvv{{\mathbf{v}}}

\DeclareMathAlphabet{\mathsfit}{\encodingdefault}{\sfdefault}{m}{sl}
\SetMathAlphabet{\mathsfit}{bold}{\encodingdefault}{\sfdefault}{bx}{n}

\usepackage[switch]{lineno}

\newcommand\blfootnote[1]{%
  \begingroup
  \renewcommand\thefootnote{}\footnote{#1}%
  \addtocounter{footnote}{-1}%
  \endgroup
}

\title{Frivolous Units: Wider Networks Are Not Really \emph{That} Wide}
\author{
  Stephen Casper,$^{^* 1,2}$ Xavier Boix,$^{^* 1,2,3}$\blfootnote{Equal contribution} \\ \textbf{Vanessa D'Amario,$^{3}$ 
  Ling Guo,$^4$ Martin Schrimpf,$^{2,3}$ Kasper Vinken,$^{1,2}$ Gabriel Kreiman$^{1,2}$}\\
}
\affiliations{
     $^1$Boston Children’s Hospital, Harvard Medical School, USA \\$^2$Center for Brains, Minds, and Machines (CBMM) \\$^3$Department of Brain and Cognitive Sciences, Massachusetts Institute of Technology, USA\\
  $^4$Neuroscience Graduate Program, University of California San Francisco, USA\\
  scasper@college.harvard.edu, xboix@mit.edu
}

\begin{document}

\newcommand*{\eg}{\emph{e.g.~}}
\newcommand*{\ie}{\emph{ie.~}}
\newcommand*{\etal}{\emph{et al.~}}
\newcommand*{\etc}{\emph{etc.}}

\newcommand{\myfootnote}[1]{
    \renewcommand{\thefootnote}{}
    \footnotetext{\scriptsize#1}
    \renewcommand{\thefootnote}{\arabic{footnote}}
}

\maketitle

\begin{abstract}

A remarkable characteristic of overparameterized deep neural networks (DNNs) is that their accuracy does not degrade when the network width is increased. Recent evidence suggests that developing compressible representations allows the complexity of large networks to be adjusted for the learning task at hand. However, these representations are poorly understood. A promising strand of research inspired from biology involves studying representations at the unit level as it offers a more granular interpretation of the neural mechanisms. In order to better understand what facilitates increases in width without decreases in accuracy, we ask: Are there mechanisms at the unit level by which networks control their effective complexity? If so, how do these depend on the architecture, dataset, and hyperparameters? 

We identify two distinct types of ``frivolous'' units that proliferate when the network's width increases: prunable units which can be dropped out of the network without significant change to the output and redundant units whose activities can be expressed as a linear combination of others. These units imply complexity constraints as the function the network computes could be expressed without them. We also identify how the development of these units can be influenced by architecture and a number of training factors. Together, these results help to explain why the accuracy of DNNs does not degrade when width is increased and highlight the importance of frivolous units toward understanding implicit regularization in DNNs.

\end{abstract}

\section{Introduction}

\begin{figure*}[t!]
\centering
\begin{tabular}{@{\hspace{-0.2cm}}c@{\hspace{-0.1cm}}c@{\hspace{-0.1cm}}c}
    \includegraphics[width=.33\linewidth]{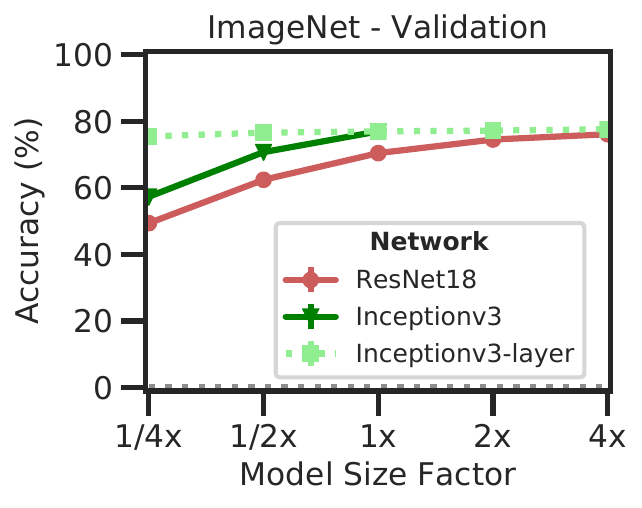}&
    \includegraphics[width=.33\linewidth]{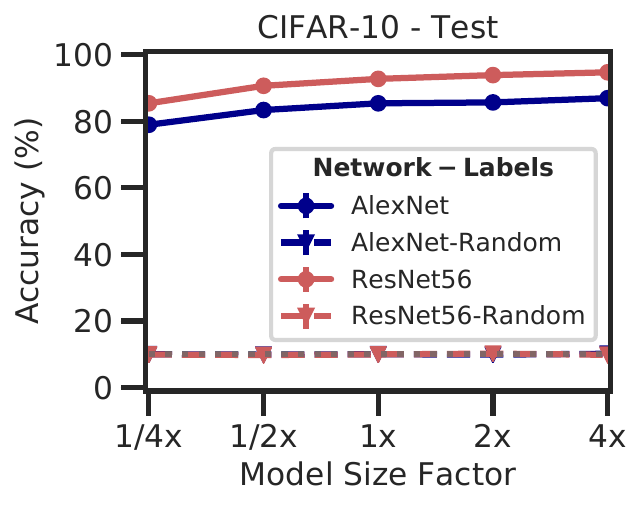}&
	\includegraphics[width=.33\linewidth]{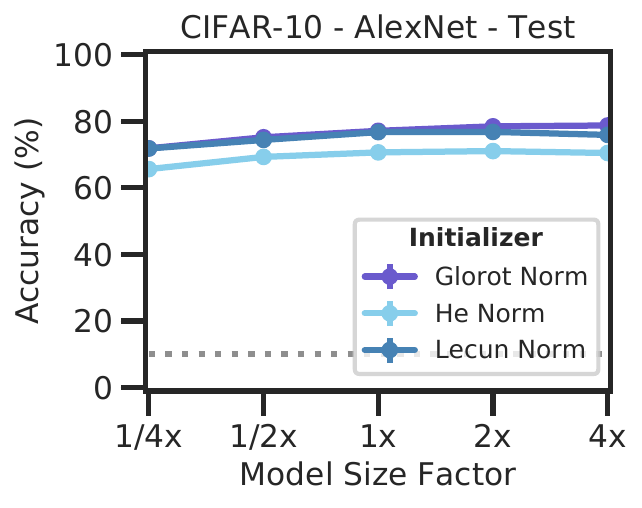}\\
	(a) & (b) & (c) \\
\end{tabular}
\caption{Test accuracy does not degrade with model width in overparameterized deep networks. Top-1 test accuracies across model sizes and datasets. (a) ResNet18s, Inception-v3s, and Inception-v3-layer (with a single layer's width varied) trained in ImageNet. (b) Regularized AlexNets and ResNet56s with and without training on random labels in CIFAR-10. (c) Unregularized Alexnet with Glorot, He, or LeCun initialization, in CIFAR-10.}
\label{fig:testaccuracy}
\end{figure*}

A striking feature of deep neural networks (DNNs) is that wider networks with more units in each layer tend to generalize as well or better than thinner ones, even when trained without explicit regularization. In practice, these networks are typically overparameterized,~\ie the number of free parameters is often several orders of magnitude greater than the number of training examples, yet wide versions avoid overfitting relative to thinner ones~\cite{neyshabur2017exploring, neyshabur2018role, Novak2018, poggio2017theory}. This phenomenon is demonstrated in Fig.~\ref{fig:testaccuracy}, in which we plot the testing accuracy of common overparameterized architectures while varying each network's width (see section~\ref{sec_methods} for details). We vary the number of units in fully connected layers and filters in convolutional layers by a size factor. The test errors of wider, more overparameterized networks do not degrade relative to thinner ones, despite a greater potential for overfitting.

\citet{frankle2018lottery} find that in certain DNNs, the crucial computations are performed by weight-sparse subnetworks with initializations primed for the learning task,~\ie such subnetworks have won the ``initialization lottery.'' They suggest that wide networks may perform as well as or better than thin ones because they ``buy more lottery tickets'' and more reliably contain these fortuitously initialized subnetworks. However, regarding subnetworks that are not part of a winning ticket, it remains unclear why they do not have a harmful effect on test performance. Some clues may come from recent works involving compression-based performance bounds~\cite{arora2018stronger, zhou2018non, suzuki2019compression}, suggesting a link between compressibility and non-overfitting. This leads to our central question: what neural mechanisms facilitate increases in network width without decreases in accuracy?


A promising strand of research inspired from neuroscience has aimed to understand network representations at the individual unit level~\cite{zeiler2014visualizing, zhou2018revisiting}. Units have been referred to as the ``building blocks of interpretability''~\cite{olah2018building}, as analyzing them allows for simple interpretations of DNNs. In this paper, we investigate the ways in which unit-level properties evolve as network width increases. Several signs point to compressible units. One example is the success of dropout~\cite{srivastava2014dropout}. Other clues come from biological brains which develop redundancies and are remarkably robust to neuronal death, suggesting that many neurons are not necessary for short-term performance~\cite{strehler1980randomness, glassman1987hypothesis}.

Here, we identify two types of \emph{frivolous} units which emerge in greater proportions when the network's width is increased: \emph{prunable units} which can be dropped out of the network without significant change to the output, and \emph{redundant units} whose activities can be expressed as a linear combination of others in the same layer. These units are complexity constraints because the function the network represents could be expressed by a thinner network without them. We show that the rate at which these frivolous units appear consistently outpaces the growth of the network as a whole. This suggests that they play a major role in how DNNs constrain their complexity as width is increased. Furthermore, these results add to our understanding of the effects of \emph{implicit regularization} by showing that frivolous units help a network adapt to the complexity of the task at hand.



\section{Related Work}

\noindent{\textbf{\emph{Understanding Implicit Regularization.}}} DNNs exhibit fascinating properties related to generalization including double descent~\cite{mei2019generalization, nakkiran2019deep, d2020double} and the ability memorize entire datasets with random labels while still generalizing well when trained on uncorrupted data~\cite{Zhang2016}. A number of works have aimed to uncover causal mechanisms that explain why, out of the many optima DNNs can reach, they tend toward simple and geometrically smooth solutions that generalize~\cite{kubo2019implicit, neyshabur2017exploring, neyshabur2018role, Zhang2017b, poggio2017theory, Novak2018}. In particular, some have shown that DNNs fit simple patterns more readily than complex ones~\cite{ansuini2019intrinsic, arpit2017closer, gidel2019implicit, derandom}, that deep ReLU networks tend to exhibit ``surprisingly few'' activation patterns~\cite{hanin2019deep, xiong2020number}, and that the intrinsic dimension of common learning tasks is much lower than the number of network parameters~\cite{li2018measuring}. 

Instead of focusing on the broader question of why DNNs generalize, this work relates to the subproblem of understanding how performance does not degrade with increases in width. This question follows in part from the lottery ticket hypothesis~\cite{frankle2018lottery} as it remains unclear how the capacity of wider networks does not degrade accuracy.


\begin{figure*}[t!]
\centering
\begin{tabular}{@{\hspace{-0.2cm}}c@{\hspace{-0.05cm}}c@{\hspace{-0.05cm}}c@{\hspace{-0.05cm}}c}
    \includegraphics[width=.25\linewidth]{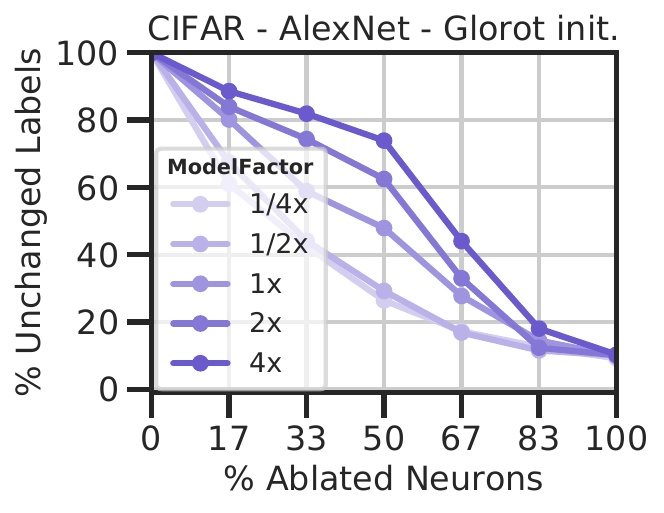} &
    \includegraphics[width=.25\linewidth]{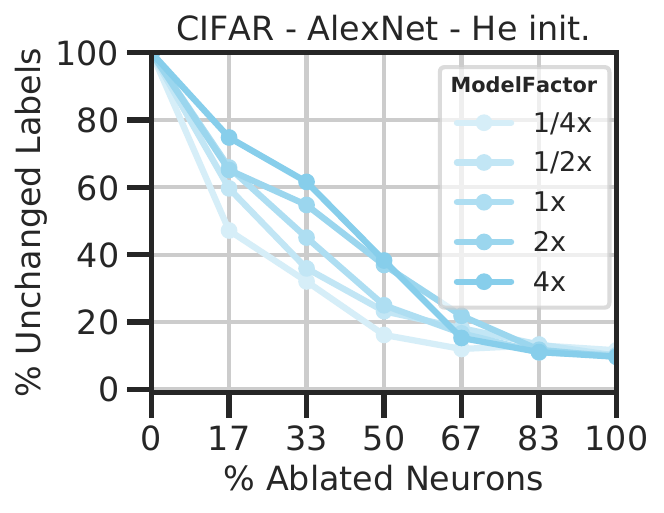} & 
    \includegraphics[width=.25\linewidth]{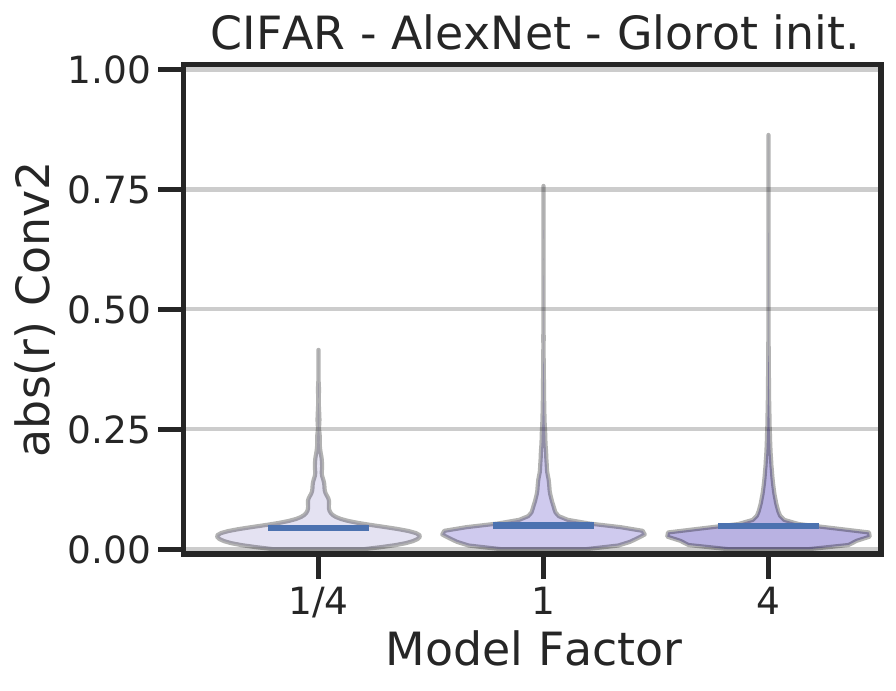} &
	\includegraphics[width=.25\linewidth]{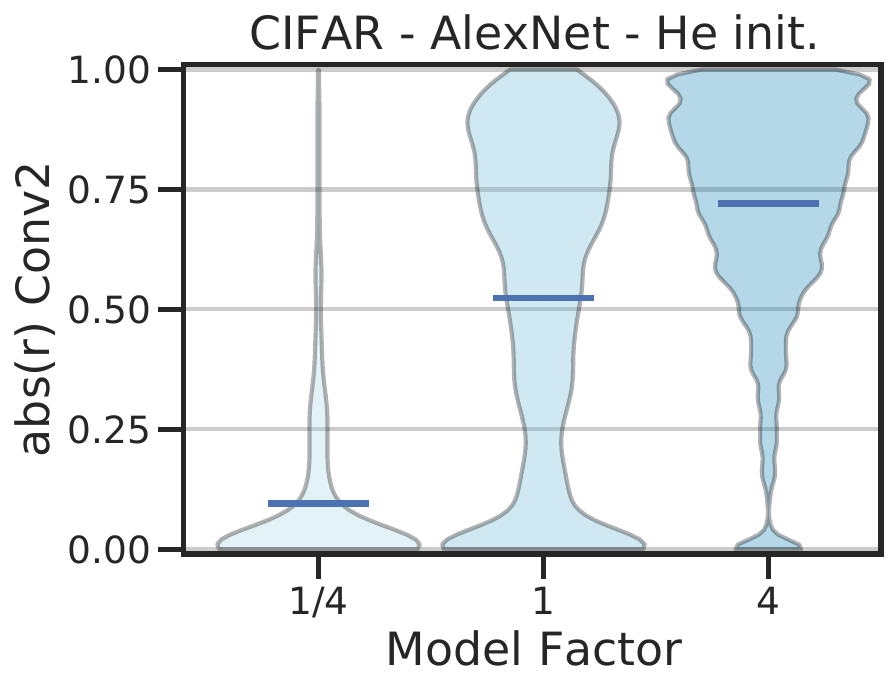} \\
	(a) & (b) & (c) & (d) \\ 
\end{tabular}
\caption{Networks develop units which are prunable and units that are linearly dependent to varying degrees. All networks were identically trained AlexNets except for width and  initialization. No explicit regularization was used. Glorot (low-variance) initialized networks are in purple, and He (high-variance) in blue. (a-b) Prunable units: points show the proportion of labels that did not change when the percentage of units on the x-axis was randomly ablated. (c-d) Redundant units: violin plots with absolute-valued correlation coefficients between unit activations for the final convolutional layer across the testing set.}
\label{fig:concept_examples}
\vspace*{-0.35cm}
\end{figure*}


\noindent{\textbf{\emph{Understanding Networks at the Unit Level.}}} A promising framework inspired from neuroscience is understanding representations at the unit level because it leads to elementary and intuitive interpretations of the neural mechanisms. In artificial DNNs, these methods tend to focus on analyzing how units respond to data~\citep{zeiler2014visualizing, montavon2018methods}. Much progress has been made in developing semantic interpretations by optimizing inputs to maximize the activations of units \cite{olah2017feature, olah2018building}, ablational analysis both in recognition networks and GANs~\cite{zhou2018revisiting,bau2018gan}, and interpreting units via a dataset of visual semantic concepts~\cite{bau2017network,mu2020compositional}. 


\noindent{\textbf{\emph{Network Compression.}}} Recent evidence suggests that networks develop compressible representations to adjust their complexity to the learning task at hand~\cite{arora2018stronger, zhou2018non, suzuki2019compression}. Broadly speaking, existing compression algorithms fall into four distinct categories: quantization, knowledge distillation, parameter pruning, and low-rank factorization~\cite{cheng2017survey, choudhary2020comprehensive}. Of these, we focus on pruning and low-rank factorization because they both allow for a simple mapping from an overparameterized network to a compressed one with fewer parameters. Several works have studied compression by pruning low-valued weights and have shown that the number of parameters can often be compressed by an order of magnitude or more this way~\cite{fukuyama2014state, molchanov2017variational, tung2018clip, ma2019resnet}. However, weights are more difficult to interpret than units, and these methods prune weights based on magnitude alone without analyzing how the network behaves across a dataset. Contrastingly, the approaches we discuss in the following section are data-driven and unit-centric.


\section{Frivolous Units} \label{sec_hypothesis}

Here, we introduce unit-level mechanisms that facilitate increases in network width without decreases in accuracy. We demonstrate that these lead to a function computed by the wider network that can be well-approximated by a thinner one and hence help to explain how performance does not degrade with width. Let $\mathcal{N}$ and $\mathcal{W}$ denote respectively a narrow and a wide DNN with the same architecture which represent mappings from datapoints to labels denoted as $f_\mathcal{N}$ and $f_\mathcal{W}$. Thus, to explain how $\mathcal{W}$ does not utilize its full capacity, we show that $f_\mathcal{W}$ can be expressed with a thinner network,~\ie $f_\mathcal{W} \approx f_\mathcal{N}$ with respect to a data distribution.

To investigate what representations $\mathcal{W}$ develops to facilitate that $f_\mathcal{W} \approx f_\mathcal{N}$, we introduce prunable units which can be removed from a network by excision and redundant units which can be removed by factorizing their layers. Furthermore, we show that these units are distinct and can emerge independently of one another. Similar motifs have been targeted by previous compression works which focus on finding smaller networks that have similar accuracy to an uncompressed one. In contrast, we focus on showing that smaller networks exist which compute similar \emph{functions} as a larger network,~\ie $f_\mathcal{W} \approx f_\mathcal{N}$ across a data distribution. This is more stringent than evaluating accuracy, as preserving accuracy is necessary for showing $f_\mathcal{W} \approx f_\mathcal{N}$ but not sufficient. 


\noindent \textbf{\emph{Prunable Units.} }
Several algorithms based on pruning have been used to compress networks to a fraction of the original size while maintaining test accuracy \cite{han2015deep, han2015learning, hu2016network, luo2017thinet}. The fact that DNNs are robust to the removal of certain units suggests that they compute functions that have a lower complexity than their architectures are capable of.


Suppose that a narrow network, $\mathcal{N}$, is identical to a wide one, $\mathcal{W}$, but with a set of units removed and that $f_\mathcal{W} \approx f_\mathcal{N}$ across a data distribution. If so, then we refer to the removed units as \emph{prunable}. Note that, due to interactive effects from multiple units, joint and individual prunability are distinct. A variety of phenomena could result in prunability ranging from simple explanations such as units being sparsely activated or their outgoing weights being small, to more complex ones such as subsequent layers discarding their activity.

To evaluate how well removing a set of units preserves network function, we analyze the proportion of examples in a testing set whose labels do not change when units are ablated. Finding the largest set of units that can be removed from a network in a way that results in a given proportion of unchanged labels is NP-hard. Instead of searching for optimal prunable sets, to scalably measure how prunable units are on average, we analyze the proportion of unchanged labels when random ablations are applied to various numbers of units (see Section~\ref{subsec:measures} for details). Fig.~\ref{fig:concept_examples}a-b shows these results. As more units are removed, more output labels change for all networks, however, they exhibit different trends. Fig.~\ref{fig:concept_examples}a shows a case in which the labels output by wider networks are more resistant to random ablations than thinner networks. This indicates that in this case, much of the wider networks' excess capacity goes toward prunable units. Contrastingly, Fig.~\ref{fig:concept_examples}b shows a case in which there is significantly less increase in robustness in a set of networks that only differ from those in Fig.~\ref{fig:concept_examples}a by how they were initialized. However, both types of networks maintain their performance when width is increased as shown in Fig.~\ref{fig:testaccuracy}, suggesting that the networks in Fig.~\ref{fig:concept_examples}b may be developing different capacity constraints. This motivates the search for a second mechanism by which complexity constraints can be understood at the unit level.

\begin{table*}
\footnotesize
\centering
\begin{tabular}{|llllll|}
\hline
\cellcolor{Gray}\textbf{Dataset} & \cellcolor{Gray}\textbf{Network} &
\cellcolor{Gray}\textbf{Initialization} &
\cellcolor{Gray}\textbf{Optimizer} &
\cellcolor{Gray}\textbf{Regularizers} &
\cellcolor{Gray}\textbf{L.Rate-B.Size} \\\hline

Uncorr. 10 dim &  MLP & Normal$\star$ & Momentum & None & Best  \\\hline
\multirow{2}{*}{Uncorr. 10k dim} &  \multirow{2}{*}{MLP} & \multirow{2}{*}{Normal$\star$} & Momentum, SGD& \multirow{2}{*}{None} & \multirow{2}{*}{Best}  \\
& & & Adam $\star$ & & \\\hline
\multirow{1}{*}{CIFAR-10} & AlexNet & Glorot/LeCun/He$\star$ & Momentum & None, DA, DO, WD$\star$ & Best$\star$  \\
\multirow{1}{*}{(+ rand. labels$\star$)} & ResNet56 & Glorot & Momentum & BN, DA, WD & Best$\star$ \\\hline
\multirow{2}{*}{ImageNet} & ResNet18 & Glorot & Momentum & BN, DA, WD& Best$\star$   \\
& Inception-v3 & Normal & RMSProp & BN, DA, WD  & Best  \\\hline

\end{tabular}
\caption{Network training and performance details: ``BN'' refers to batch normalization, ``DA'' refers to data augmentation, ``DO'' refers to dropout, and ``WD'' refers to L2 weight decay. ``Best'' refers to learning rate/batch size combinations that achieved the highest accuracy. Stars ($\star$) indicate factors for which we tested multiple hyperparameters/variants.}
\label{tbl:nets}
\end{table*}
\normalsize

\noindent \textbf{\emph{Redundant Units.}} In contrast to compression algorithms based on pruning are ones which focus on low-rank factorization. These methods in practice have successfully compressed networks to a fraction of their original size while maintaining testing accuracy~\cite{sainath2013low, denton2014exploiting, srinivas2015data}. Most previous works using these methods compress units by representing them in weight space, but units can similarly be represented in activation space across a dataset. 
We denote as \emph{non-redundant} the largest set of units whose activations are linearly independent (though in experiments we relax this via PCA). The number of non-redundant units is equal to the rank of a matrix that represents the activations of the units across a dataset. The remaining \emph{redundant} units help to regulate the complexity of the network because a thinner network can be constructed without them such that $f_\mathcal{W} \approx f_\mathcal{N}$ across the dataset. In the Appendix, we provide algorithmic details for removing the redundant units and refactoring the outgoing weights of the non-redundant ones.

As with prunability, networks sometimes develop very different levels of redundancy at different widths. Fig.~\ref{fig:concept_examples}c-d depicts the distributions of absolute-valued unit-to-unit correlation coefficients for the final convolutional layers in two classes of networks which only differ in how they were initialized. Fig.~\ref{fig:concept_examples}c shows a case in which the level of unit-to-unit correlation increases only slightly as network width is increased. However, Fig.~\ref{fig:concept_examples}d shows a case in which much more correlation develops with wider networks. Correspondingly, these networks develop more redundant units (see Section~\ref{subsec:measures} for further details). This implies that here, excess capacity is largely utilized to form these redundant units.

\subsection{Relating Prunable and Redundant Units} \label{subsec_relating}

As reflected in Fig.~\ref{fig:concept_examples}, we clarify here that prunability and redundancy are distinct and prove by construction that they can emerge either together or independently. 
Recall that $\mathcal{N}$ and $\mathcal{W}$ denote respectively a narrow and a wide DNN with the same architecture and that each represents a mapping from datapoints to labels denoted by $f_\mathcal{N}$ and $f_\mathcal{W}$. Let $\rvu$ refer to the activity of the layer before the output layer in $\mathcal{N}$ and $\boldsymbol\theta$ the incoming weights for the output layer of $\mathcal{N}$. Thus, the output of $\mathcal{N}$ is equal to $\boldsymbol\theta^T\rvu$. Examples can be constructed of a wide network $\mathcal{W}$, twice the size of $\mathcal{N}$, such that $ f_\mathcal{N}=f_\mathcal{W}$ (not only approximately, but exactly) yet each $\mathcal{W}$ has different levels of prunability and redundancy. Note that there are other possible cases aside from the prototypes presented here.

\noindent \textbf{\emph{More of Both Prunable and Redundant Units.}} We can build a $\mathcal{W}$ to be more prunable and redundant by first duplicating the final layer of $\mathcal{N}$ such that the layer before the output has activity $[\rvu, \rvu]$ and setting the output weights equal to $[{\boldsymbol\theta},\boldsymbol{0}]$. Because $[{\boldsymbol\theta},\boldsymbol{0}]^T[\rvu, \rvu]=\boldsymbol\theta^T\rvu$, the outputs will be identical to $\mathcal{N}$. All other layers of $\mathcal{W}$ can then be constructed analogously in the order of deeper to shallower layers. In this case, both prunable and redundant units would be greater in quantity in $\mathcal{W}$ because half of the units in the network are redundant with the other half and can be pruned due to having $\mathbf{0}$-valued outgoing weights. 

\noindent \textbf{\emph{Only More Prunable Units.}} A $\mathcal{W}$ that is more prunable but not more redundant can be constructed by first making the units before the output layer in $\mathcal{W}$ equal to $[\rvu, \rvv]$, where $\rvv$ gives units with activities orthogonal to $\rvu$ and each other across the data distribution. Thus, the units are not redundant. To make $f_\mathcal{W}$ equivalent to $f_\mathcal{N}$, the outgoing weights of this layer can be set equal to $[\boldsymbol\theta,\boldsymbol{0}]$, and as in the previous case, the units that are multiplied by $0$ (half of the layer) are prunable. Then constructing other layers analogously from deeper to shallower layers results in a $\mathcal{W}$ that is only more prunable.

\noindent \textbf{\emph{Only More Redundant Units.}} We can duplicate the final layer of $\mathcal{N}$ so that $\mathcal{W}$'s final layer has activations equal to $[\rvu, \rvu]$, making half of the units redundant. Then we can obtain a network that computes the same function as $\mathcal{N}$ via ``weight balancing'' with weights leading into the output layer of $\frac{1}{2}[\boldsymbol\theta + b, \boldsymbol\theta - b]$, where $b$ is a large constant. Although $\frac{1}{2}[\boldsymbol\theta + b,\boldsymbol\theta - b]^T[\rvu, \rvu] = \boldsymbol\theta^T\rvu$, when a unit is pruned, the layer's output changes by an offset of $\frac{1}{2}(b - \theta_k)u_k$, where $u_k$ is the removed unit activation, making it not prunable when $b$ is sufficiently large. Then $\mathcal{W}$ can be completed to be only more redundant by performing the same procedure on on the rest of the network from deeper to shallower layers.

\noindent \textbf{\emph{Upshot.}} Finally, it is helpful to note that $f_\mathcal{W} \approx f_\mathcal{N}$ does not necessarily imply that frivolous units {must} exist in $\mathcal{W}$ because capacity constraints need not emerge at the unit level, and circuits of different complexity can still compute similar functions. We provide clarifying examples in the Appendix. Also in the Appendix, we show that for randomly initialized networks, prunability and redundancy will, in expectation, be proportional to the network width. However, it remains unclear precisely how they emerge in trained networks. As we will show in the following sections, the growth of frivolous units tends to outpace the growth of a network as a whole. 

\section{Methods}\label{sec_methods}

\begin{figure*}[]
\centering
\begin{tabular}{@{\hspace{-0.2cm}}c@{\hspace{-0.1cm}}c@{\hspace{-0.1cm}}c@{\hspace{-0.1cm}}c}
  \includegraphics[width=.26\linewidth]{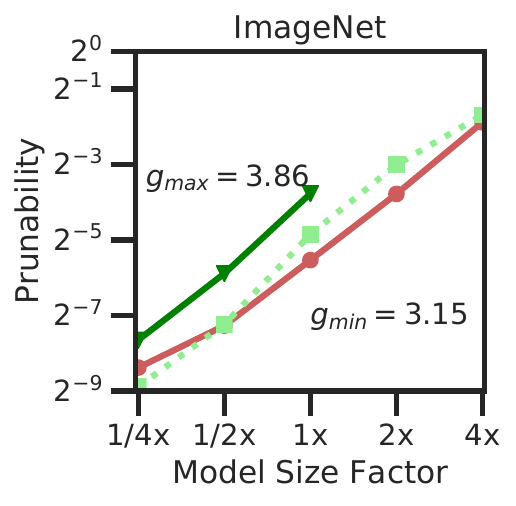} &
  \includegraphics[width=.26\linewidth]{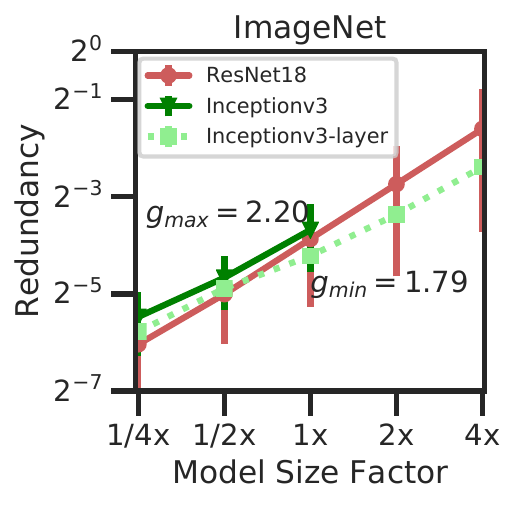} &
    \includegraphics[width=.26\linewidth]{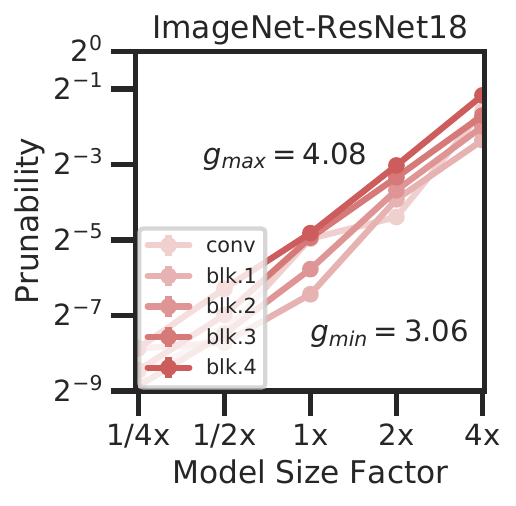} &
    \includegraphics[width=.26\linewidth]{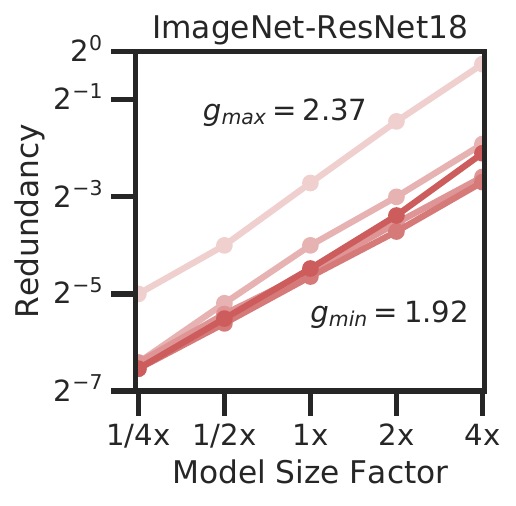}\\
	(a) & (b) & (c) & (d) \\
\end{tabular}
\caption{Prunability strongly outpaces the growth of the network as a whole and redundancy increases at rates similar to the network in ResNet18s and Inception-v3s. (a) Prunability. (b) Redundancy. (c) Prunability layerwise for ResNet18s. (d) Redundancy layerwise for ResNet18s. The gain, $g$, represents the increase in prunability or redundancy when the network size is doubled. Max and min $g$ values are given for each plot.}
\label{fig:imagenet_results}
\vspace{-0.3cm}
\end{figure*}

Table~\ref{tbl:nets} gives training details. Features we tested multiple variants of are marked with a star ($\star$). Further details for all networks are in the Appendix. To see how  prunability and redundancy vary with width, we tested network variants in which the number of weights/filters in each layer/block/module were multiplied by factors of $1/4$, $1/2$, $1$, $2$, and $4$. 

\noindent \textbf{\emph{Datasets.}} For larger scale experiments, we used the ImageNet \cite{ILSVRC15} and CIFAR-10 \cite{krizhevsky2009learning} datasets. For small-scale experiments, we used train and test datasets of 1,000 binary-labeled examples generated by $1/4$x-sized randomly-initialized multilayer perceptrons (MLPs) from uncorrelated Gaussian inputs. We verified these labeling MLPs to output each label for between 40\% and 60\% of inputs. We present results for the test sets, but they were nearly identical for the train sets. 

\noindent \textbf{\emph{Networks.}} For ImageNet, we used ResNet18s from \citet{he2016deep} and Inception-v3 networks from \citet{szegedy2016rethinking}. Due to hardware limitations (we used a dgx1 with 8x NVIDIA V100 GPUs 32GB), we could not train any $2$x or $4$x Inception-v3s and instead experimented with versions with a single layer varied from $1/4$x to $4$x (denoted as \emph{Inception-v3-layer} in plots). For experiments with CIFAR-10, we used AlexNet models based on \citet{Zhang2016} and ResNet56s from \citet{he2016deep}. For small-scale experiments, we used simple MLPs with 1 hidden layer of 128 units for the 1x size. For all networks, increasing model size resulted in equal or improved performance as shown in Fig. \ref{fig:testaccuracy}. In the Appendix, we plot the number of trainable parameters for each network.

\subsection{Measuring Prunability and Redundancy} \label{subsec:measures}

We measure frivolous units via ablation and linear analysis of activations which allows us to experiment with hundreds of networks including convolutional nets at the ImageNet scale. 

\noindent \textbf{\emph{Prunability.} }
To measure how prunable units are on average, we analyze robustness to random ablation. We first find what proportion of labels for the test set do not change when varying proportions of units are ablated. These are applied to fully connected layers and feature map outputs in convolutional layers such that each spatial location is treated as a different unit. After obtaining ablation curves as shown in Fig.~\ref{fig:concept_examples}, we use linear interpolation to estimate the proportion of units which can be randomly ablated on average with a tolerance for label corruption of $0.2$ (though trends were similar for different tolerance levels which we tested up to $0.5$). This procedure is repeated three times with randomly chosen units and the set that yields the largest number of prunable units is selected. Finally, we divide by the number of units in the 4x model to normalize the results.

\noindent \textbf{\emph{Redundancy.}} To quantify redundancy, we collect each layer's activation matrix across the test set. In contrast with our method for prunability, for convolutional layers, we treat each feature map output as a unit and consider different spatial locations to be examples for the same unit. We do this because (1) outputs of the same feature map are similar in the sense that they can have the same activity by shifting the image and (2) for computational tractability. We use principal component analysis on the activations to calculate the number of units that are redundant with a tolerance of 0.05 for the proportion of unexplained variance (though trends were similar for different tolerance levels which we tested up to 0.15). We divide by the size of that layer in the 4x model to normalize the results and report the average across all layers of a network weighting each equally regardless of size.

\section{Results}

\label{results_section}
\label{sec_results}

\begin{figure*}[t!]
\centering
\begin{tabular}{@{\hspace{-0.2cm}}c@{\hspace{-0.1cm}}c@{\hspace{-0.1cm}}c@{\hspace{-0.1cm}}c}

    \includegraphics[width=.26\linewidth,valign=T]{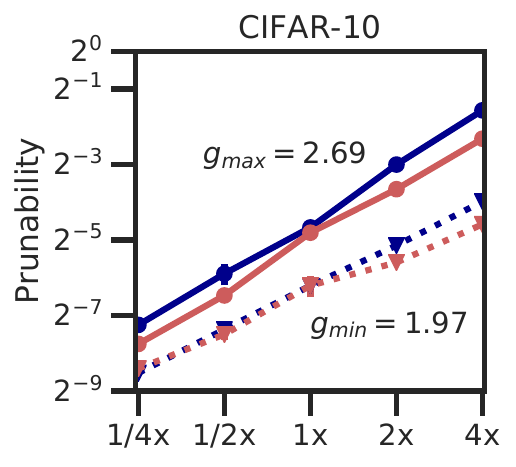} & 
	\includegraphics[width=.26\linewidth,valign=T]{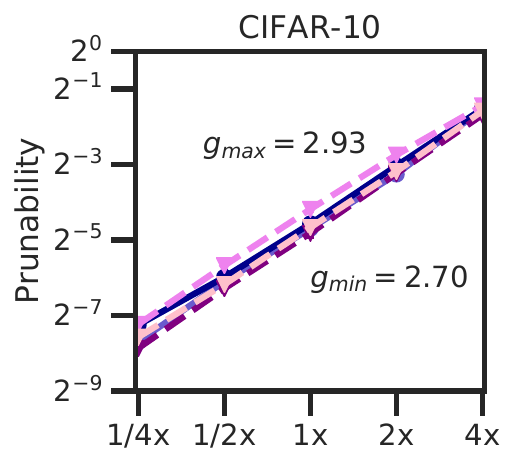}&
	\includegraphics[width=.26\linewidth,valign=T]{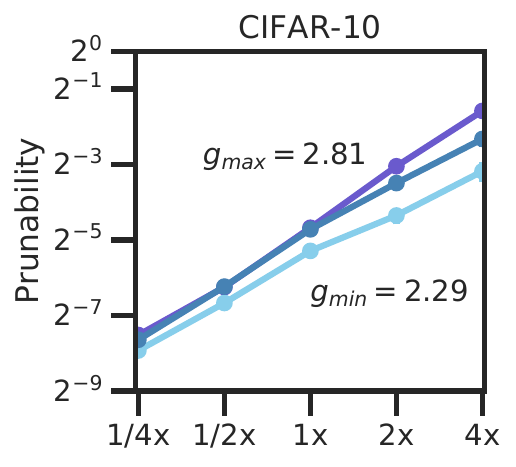} &
	\includegraphics[width=.26\linewidth,valign=T]{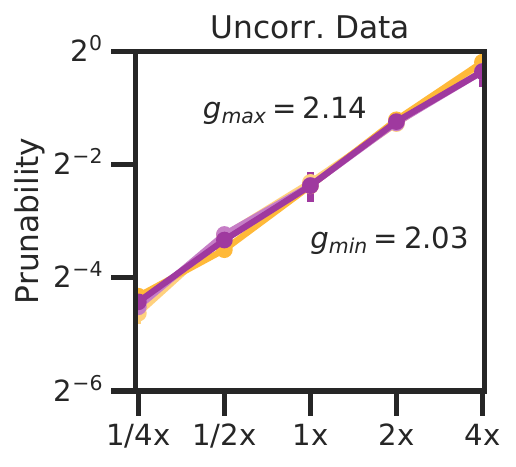}\\

    \includegraphics[width=.26\linewidth]{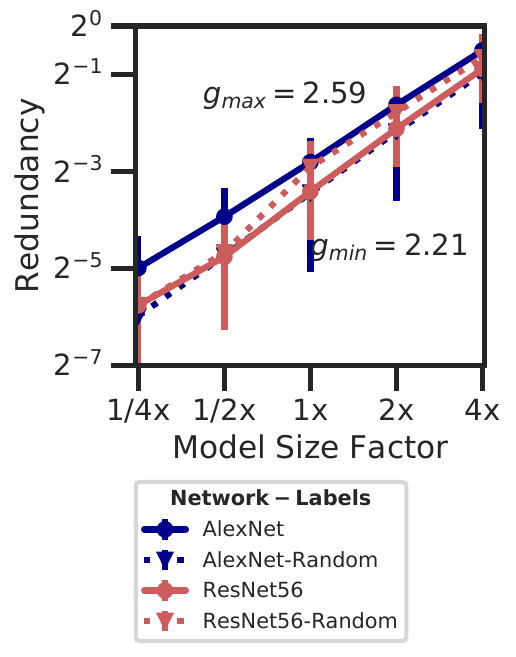} & 
	\includegraphics[width=.26\linewidth]{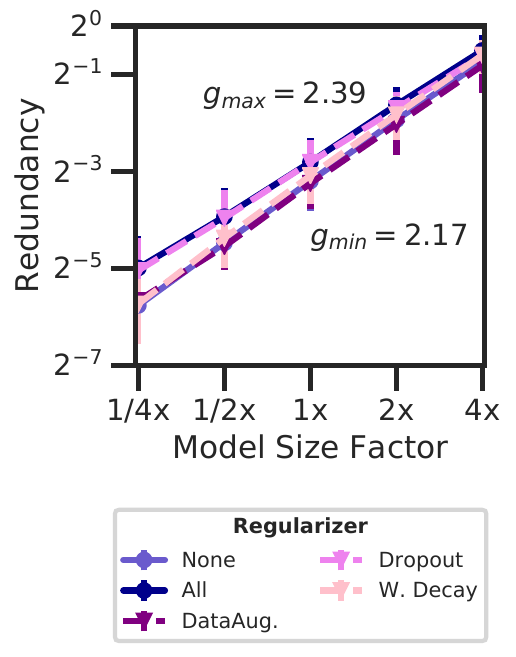}&
	\includegraphics[width=.26\linewidth]{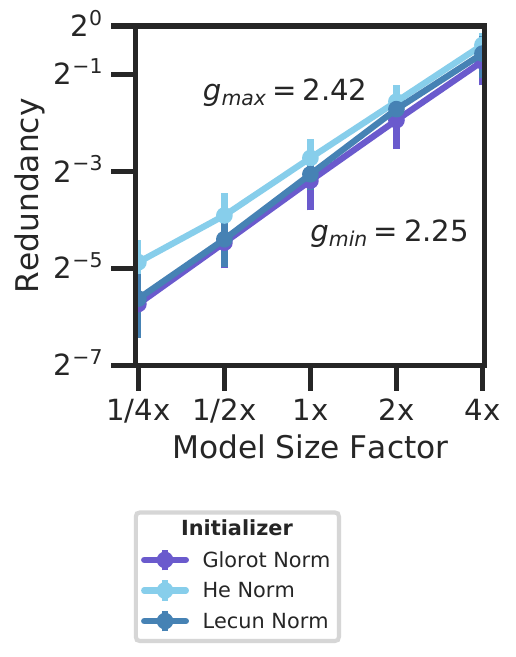} &
	\includegraphics[width=.26\linewidth]{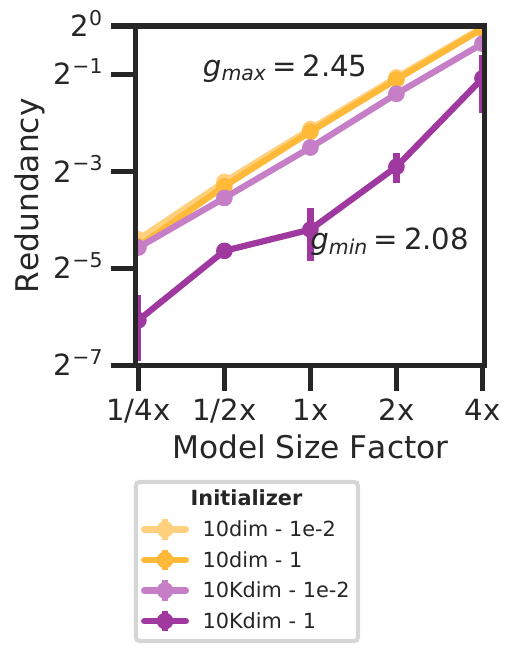}\\
	
	(a) & (b)& (c)& (d)\\
	
\end{tabular}
\caption{Prunability and/or redundancy outpace the growth of the overall network for different architectures, regularizers, initializers and datasets. (a) AlexNets and ResNet56s trained with and without random labels (CIFAR-10). (b) AlexNets trained with and without regularization (CIFAR-10). (c) AlexNets trained with various initializations (CIFAR-10). (d) MLPs with different trainsets and initializations (synthetic data). Max and min gain factors are given for each plot.}
\label{fig:results_other}
\vspace*{-0.4cm}
\end{figure*}

Here, we introduce how prunability and redundancy vary as a function of model width across size factors from 1/4x to 4x. We use a $\log$ scale for both axes in all plots and include maximum and minimum ``gain'' values for the curves in each plot represented as $g_{\min}$ and $g_{\max}$. This gain value gives the average increase in the number of prunable and redundant units as network width is doubled. A gain $g > 2$ indicates that a type of unit more-than-doubles on average when the network width is doubled. Excluding Figs.~\ref{fig:imagenet_results}a,c,d, points are averages across three trials. Error bars giving standard deviations are provided, but do not always appear at the given scale. See the Appendix for further details.

\noindent \textbf{{Prunability and/or redundancy increase at a rate greater than units of the network overall.}} Fig.~\ref{fig:imagenet_results}a-b and Fig.~\ref{fig:results_other}a-d show that both types of units consistently emerge across experiments. For all but two cases (redundancy in Inception-v3s and prunability in ResNet56s trained on random labels), $g_{\min} > 2$ meaning that the growth rate of frivolous units is greater than that of the overall network. This demonstrates implicit regularization at the unit level by showing that in wider networks, the distribution of units tends to prefer frivolous ones more than in thinner networks. We also plot trends for the individual layers of ResNet18s in Fig.~\ref{fig:imagenet_results}c-d and Inception-v3s in the Appendix. While frivolous units consistently emerge in individual layers when their width is increased, they do so to varying extents, and we find no consistent relationship between depth and frivolity. 

The fact that frivolous units tend to more-than-double is mirrored by the fact that their non-frivolous ones tend to less-than-double which is shown in the Appendix. We also find that although non-prunable and non-redundant units increase at a lower rate, they increase nonetheless. It is notable that even if the rate of increase is larger for frivolous units, if a network has a small proportion of them to begin with, it can develop more new non-frivolous units than new frivolous ones as width increases. 
This suggests avenues for future work toward understanding non-prunable and non-redundant units using methods other than analysis of prunable and redundant units separately. A promising possibility is analysis of both jointly. Note that in some networks, particularly Inception-v3s and ResNet18s in Fig.~\ref{fig:imagenet_results}a-b, trends in prunability and redundancy differ significantly which demonstrates that these measurements respond to distinct sets of units.

\noindent \textbf{{Prunability and redundancy develop in networks trained on random labels.}} To better understand the relationship between frivolous units and generalization, we analyze networks trained to memorize randomly labeled images. Fig.~\ref{fig:results_other}a compares the results with AlexNet and ResNet56 models trained with and without random labels. With random labels, all networks of the 1x size or greater fit the training set with at least 99\% accuracy. Even when fitting noise, these models increase the quantity of frivolous units and even increase the number of redundant ones with $g_{\min} > 2$. This demonstrates that although the emergence of frivolous units with $g > 2$ implies implicit regularization, it does not imply generalization.

\begin{figure*}[]
\centering
\begin{tabular}{cc}
    (a) & \includegraphics[width=.92\linewidth]{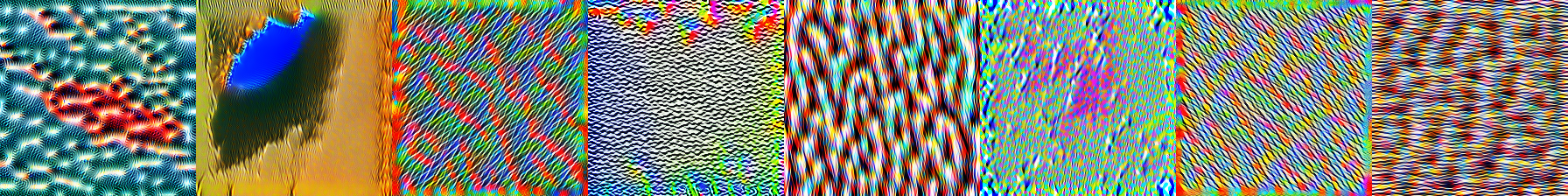} \\
     (b) & \includegraphics[width=.92\linewidth]{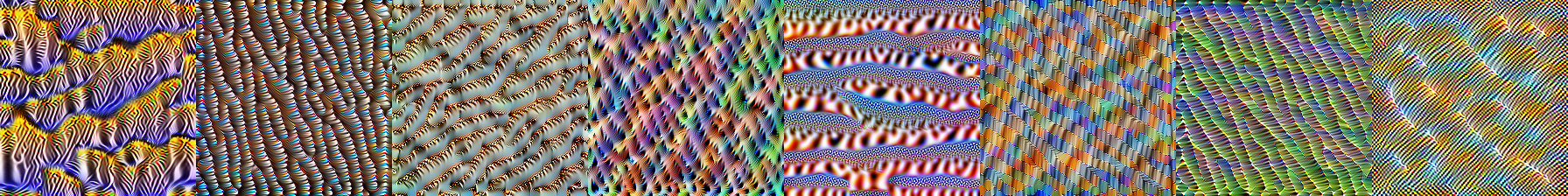} \\
\end{tabular}
\caption{Example visualizations of 8 units from the first block of the 1/4x (a) and 4x (b) ResNet-18s. The total variations of these images and their Fourier transforms were used for hypothesis testing via a rank-based permutation test.} 
\label{fig:lucid_vis_small}
\end{figure*}

\noindent \textbf{{Trends are similar under explicit regularization.}} To compare the effects of implicit and explicit regularization, we ask how explicit regularizers influence the emergence of frivolous units. In Fig.~\ref{fig:results_other}b, we show that data augmentation, dropout, weight decay, and all three together have only modest effects on the trends in frivolous units. This suggests that implicit regularization may operate at the unit level in significantly different ways than explicit regularization.

\noindent \textbf{{Initialization influences prunability and redundancy.}} Some recent works have suggested that network initialization has a large influence over generalization behavior \cite{frankle2018lottery, chizat2019lazy, woodworth2019kernel}. To analyze its effects on frivolous units, we test three common methods of weight initialization. Fig.~\ref{fig:results_other}c presents results for AlexNets trained with Glorot \citep{glorot2010understanding}, He \citep{he2015delving}, and Lecun \citep{lecun2012efficient} initializations which each initialize weights using Gaussian distributions with variances depending on the layer widths. 
In these networks, as the initializations change, there is a tradeoff between the emergence prunable and redundant units. We also display the same curves for uniform-distributed versions of these initializations in the Appendix and find the results to be similar, suggesting that the initialization distribution matters little compared to variance.

\noindent \textbf{{Datasets influence prunability and redundancy.}} To see if frivolous units result from structure in the input data, we train MLPs on uncorrelated data with labels from randomly initialized teacher networks. Fig.~\ref{fig:results_other}d, shows the results of altering initialization variance in MLPs trained on these datasets with examples of 10 and 10,000 dimensions. Results from the full hyperparameter searches are in the Appendix. Despite uncorrelated training data, prunability and redundancy emerge nonetheless with $g > 2$, though not as large as for other experiments. For the case with high initialization variance and high input dimension, the MLPs have similar $g$ values for prunability to other networks but have a much lower number of prunable units. This indicates an interaction between initialization and dataset in the emergence of prunable units. 

\noindent \textbf{{Results are consistent across optimizers, learning rates, batch sizes, and number of training epochs.}} Results from additional experiments are in the Appendix. 

\noindent \textbf{{Network width influences interpretability but with no consistent trend across layers.}} Given that prunable units do not seem to represent any features essential for the task at hand and that redundant units may be representable as a combination of multiple feature directions, we ask whether these units degrade interpretability in wide networks. To test this, we use Lucid~\cite{olah2017feature} to generate images optimized to maximize the activation of units in 1/4x and 4x ResNet18s and 1/4x and 1x Inception-v3s. Examples from the first block of ResNet-18s are shown in Fig.~\ref{fig:lucid_vis_small} (see Appendix for more visualizations). As a proxy measure for interpretability, we calculate the \emph{total variation} of these visualizations which measures how different the values of adjacent pixels are. The total variations were analyzed both for the raw visualizations and their Fourier transforms (lower total variation means better interpretability). To compare results between thin and wide networks for each block/module, we use a rank-based permutation test. These tests provide evidence that there are differences between the thin and wide networks across layers, but not that the thin ones are consistently more interpretable. Thorough details including a table of $p$ values and effect sizes are in the Appendix.

\section{Discussion and Conclusions}

We have analyzed the emergence of prunable and redundant units in relation to the fact that the generalization ability of DNNs does not tend to decrease as network width increases. Our results show that the number of prunable and/or redundant units increases at a rate which outpaces that of the network overall which suggests that complexity-constraining features in deep networks emerge largely at the unit level. Thus, we offer the following hypothesis: consider a narrow deep network $\mathcal{N}$ and a wide one $\mathcal{W}$, both with the same architecture, trained on the same data with the same regularization and initialization schemata, and each with a tuned set of hyperparameters. Alongside $\mathcal{W}$ generalizing as well or better than $\mathcal{N}$, $\mathcal{W}$ will develop an increased proportion of prunable and/or redundant units relative to $\mathcal{N}$ due to implicit regularization. We hope that future work extending or challenging this hypothesis will lead to a richer understanding of the emergent properties of DNNs.

Despite much recent progress, to our knowledge, this is the first work to date that has quantitatively studied prunability and redundancy together in context of implicit regularization. These methods which target multiple compressible features could be used to improve upon compression based bounds~\cite{arora2018stronger} including for approaches such as that of \cite{zhou2018non} which only considers prunability but not redundancy. We also reveal specific architectures, initializations, and training methods that can be used to control what types of compressible features networks develop. Nonetheless, the fact that both of these types of units consistently proliferate highlights a need for hybrid compression methods which target both.

The framework used here can also be useful for understanding questions about complexity, robustness, and redundancy in neuroscience. Given that brains are often surprisingly capable of recovering from damage and neuronal death, it is understood that they are overparameterized networks which are both highly prunable and have built-in redundancies \cite{strehler1980randomness}. In fact, \citet{glassman1987hypothesis} estimates that at least half of the neurons in the human brain fit under our definition of frivolous. The neural activity of DNNs for object recognition has been shown to resemble neural activity in parts of the brain~\cite{yamins2016using}, and several works have aimed to understand redundancy and robustness in brains using artificial networks as models~\cite{schuster2008robust, aerts2016brain, hammelman2016artificial}. Thus, the framework presented in this paper could be helpful to understand prunability and redundancy in brains and how biological networks implicitly regularize.


While we find that frivolous units are necessary to understand implicit regularization, we do not find that they are sufficient. Although non-frivolous units increase with width at a smaller rate, they increase nonetheless. Future work should investigate the causal factors behind the formation of frivolous units and expand on understanding simplicity in network representations using insights from 
network distillation \citep{hinton2015distilling}, subnetwork analysis, or 
kernel-inspired analysis~\cite{jacot2018neural}.
Nonetheless, frivolous units play a key role in how networks constrain their effective complexity and offer a milestone toward understanding them at the unit level.




\section*{Code Availability}
Code and trained models in ImageNet are available at:\\ \url{https://github.com/biolins/frivolous_dnns}

\section*{Acknowledgments}
We are grateful to Tomaso Poggio for helpful feedback
and discussions. We also thank NVIDIA for the donation of the DGX-1 used in the experiments. This work is supported by the Center for Brains, Minds and Machines (funded by NSF STC award
CCF-1231216), the Harvard office for Undergraduate Research and Fellowships, Fujitsu Laboratories (Contract No. 40008401 and 40008819) and the MIT-Sensetime Alliance on Artificial
Intelligence. 


\section*{Ethics Statement}

This work may contribute to consequential progress in three areas:
(1) \textit{Designing robust networks:} understanding how prunability and redundancy emerge allows for more control over what types of representations they develop. This may lead to insights on improving network design and training.
(2) \textit{Interpretability:} knowledge of frivolous units is helpful for interpreting networks at the unit level which can be valuable for developing a more basic understanding and for verifying properties of learning systems regarding performance or robustness. We expect this to be primarily beneficial, especially for systems in safety-critical settings. 
(3) \textit{Compression:} our findings suggest directions for work in advancing compression techniques which could improve space and time efficiency in deep learning models. This is of particular interest to mobile and web development and could make these systems significantly more accessible. However, this may come with concerns involving the reliability of these systems or their proliferation outpacing effective oversight.

We join with others in the research community calling for the safe and judicious use of AI in ways that benefit all of humanity.

\small
\bibliography{library}

\clearpage
\appendix


\renewcommand\thefigure{\thesection \arabic{figure}}    
\setcounter{figure}{0} 

\section{A Wider Network without Frivolous Units can Compute the Same Function as a Thin Network (Section 3)}

We demonstrate in section~\ref{results_section} that as we increase the width of a network, frivolous units tend to grow at a rate which outpaces the network overall. One may ask whether this is a surprising result or if frivolous units must necessarily emerge if a wider network computes a similar function to a thin one. As we show here, a wider network need not have frivolous units in order to compute a similar function to a thinner one.


For a wider network, there can be different ways to compute the same function as a thinner network which require a different minimum number of units. For example, consider a dataset of $N+1$ dimensional datapoints in which for each, the parity of the last $N$ dimensions is equal to the value of the first dimension. Then suppose that each point were associated with a label that was equal to the identity of the first component or equivalently, the parity of the final $N$. If so, this labeling function could be computed equally well by evaluating the identity of the first dimension or the parity of the last $N$. For networks with a single hidden layer, the identity method would require only one unit while the parity method could be computed with $2^N$ hidden units (each representing a miniterm that detects one of the possible inputs, as shown in~\cite{bengio2007scaling}). Thus, a wide network with $2^N$ units could correctly label the data with all units being equally non-frivolous, but also with a single unit capturing the first dimension and the rest being frivolous.

Another reason why a wider network computing a similar function to a thin network may not develop disporoportionately more frivolous units is because of capacity constraints at the weight level as opposed to the unit level. The success of weight-pruning as a common regularization and compression method (\eg \cite{fukuyama2014state, molchanov2017variational, frankle2018lottery, tung2018clip, ma2019resnet}) suggests that complexity constraints can and do appear at this level in addition to the unit level.

\section{Prunability and Redundancy in Untrained Networks (Section 3)}

Consider an untrained network in which the weights within each layer are initialized i.i.d. from some distribution. We show here that for large datasets, as the model width is increased, the expected number of prunable units and redundant units will each increase proportionally to the width. In other words, the expectation of $g$ as defined in section~\ref{results_section} will be 2 for untrained networks. 

\noindent \textbf{Prunability:} Given any fixed data distribution, the units in the final layer of a network will have a distribution of input values determined by the network's weights. By the i.i.d. initialization of each weight, each hidden unit will contribute equally in expectation to the activational variance of each unit in the final layer. Then by the fact that each layer influences the output layer only by feed-forward action, for a layer of $n$ units, each will in expectation be responsible for $1/n$'th of the variance for each output unit. Consequently, randomly ablating a proportion $p$ of the units in any layer will, in expectation, reduce the variance of each output unit by a factor of $p$ regardless of the layer's size. So robustness to the random ablation of a given \emph{proportion} of units will be constant regardless of width for untrained networks (\ie $\mathbb{E}(g)=2$).

\noindent \textbf{Redundancy:} If a large dataset of points is used, the activations of each unit will be uncorrelated due to the i.i.d. randomness of the network's weight initialization. So in expectation, redundancy will be proportional to a network's size (\ie $\mathbb{E}(g)=2$).

\section{Activational Low-Rank Factorization Algorithm (Section 3)}

Consider an $m \times n$ weight matrix $W$ for a fully-connected layer.
For a dataset of $d$ examples, let $A$ be the $d \times m$ matrix whose rows give the activations of layer $L_i$ for each example. If so, then the inputs to layer $L_{i+1}$ will be given by the matrix product $AW$.

The goal of finding a low rank refactorization of $L_i$ based on activations is to utilize a basis of $m' < m$ units and a refactored $m' \times n$ weight matrix $W'$ such that if $A'$ is the $d \times m'$ matrix giving the activation of the basis, then $A'W' \approx AW$. For any basis of $m'$ units in $L_i$, in order to achieve $A'W' \approx AW$, both sides can be multiplied by the left inverse of $A'$ denoted as $A_L^{'-1}$ to find the optimal $W' = A_L^{'-1}AW$. In order to approximate the number of units needed for such as basis, we use principal component analysis on $A$. By analyzing the eigenvalues of the covariance matrix of $A$, we calculate the minimal number of components needed to reconstruct $A$ from $A'$ with a given error tolerance on the $L2$ distance.

\section{Methodology (Section 4)}

\label{net_details}

\subsection{Network Implementations}

\noindent \textbf{ResNet18s (ImageNet):} We use off-the-shelf models and the established training procedure from \citet{he2016deep}. They consisted of an initial convolution and batch normalization followed by four building blocks (v1) layers, each with two blocks and a fully connected layer leading to a softmax output. All kernel sizes in the initial layers and block layers were of size $7 \times 7$ and stride $2$. All activations were ReLU. In the 1x-sized model, the convolutions in the initial and block layers used $64$, $64$, $128$, and $256$ filters respectively. After Glorot initialization~\cite{glorot2010understanding}, we trained them for $90$ epochs with a default batch size of $256$ and an initial default learning rate of $1$ which decayed by a factor of $10$ at epochs $30$, $60$, and $80$. Optimization was done with stochastic gradient descent using $0.9$ momentum. We used batch normalization, $0.0001$ weight decay, and data augmentation with random cropping and horizontal flipping. Results were generated using the ImageNet validation set of 50,000 images. 

\noindent \textbf{Inception-v3s (ImageNet):} We used off-the-shelf models and the established training procedure from \citet{szegedy2016rethinking} following the established training procedure. For the sake of brevity, we will omit architectural details here. After using a truncated normal initialization with $\sigma=0.1$, we trained these networks with a default batch size of $256$ and initial default learning rate of $1$ with an exponential decay of $4\%$ every $8$ epochs. Training was run for $90$ epochs on ImageNet using the RMSProp optimizer. We used a weight decay of $0.00004$, batch normalization using $0.9997$ decay on the mean and an $\epsilon$ of 0.001 to avoid dividing by zero, and augmentation using random cropping and horizontal flipping. Due to hardware constraints, we were not able to train 2x and 4x variants of the network (we used a dgx1 with 8x NVIDIA V100 GPUs 32GB). Instead, we trained the 1/4x-1x sizes along with versions of the network with 1/4x-4x sizes for the "mixed 2: 35 x 35 x 288" layer only. We generate results using the ImageNet validation set of 50,000 images. 

\noindent \textbf{AlexNet (CIFAR-10):} We use a scaled-down version of the network developed by \citet{krizhevsky2012imagenet} similar to the one used by \citet{Zhang2016} for CIFAR-10. The network consisted of 4 hidden layers: two convolutional layers with 96 and 256 kernels respectively and two dense layers with $384$, and $192$ units in the 1x model size. In each convolutional layer, $5 \times 5$ filters with stride $1$ were applied, followed by max-pooling with a $3 \times 3$ kernel and stride $2$. Local response normalization with a radius of $2$, alpha $= 2 * 10^{-5}$, beta $= 0.75$ and bias $= 1.0$ was applied after each pooling. Each layer contained bias terms, and all activations were ReLU. We used Glorot initialization~\cite{glorot2010understanding} by default and trained these networks  with early stopping based on maximum performance on the $5,000$ image CIFAR-10 validation set.  Weights were optimized with stochastic gradient descent using $0.9$ momentum with an initial learning rate of $0.01$, exponentially decaying by $5\%$ every epoch. By default, we used a batch size of $128$, and no explicit regularizers. We generate results using the 10,000 image testing set.

\noindent \textbf{ResNet56 (CIFAR-10):} These networks were used off-the-shelf from \citet{he2016deep}. They consisted of an initial convolution and batch normalization followed by three building block (v1) layers, each with nine blocks, and a fully connected layer leading to a softmax output. In the 1x-sized model, the convolutions in the initial and block layers used $16$, $16$, $32$, $64$, and $128$ kernels respectively. Kernels in the initial layers and block layers were of size $3 \times 3$ and stride $1$. All activations were ReLU. After Glorot initialization~\cite{glorot2010understanding}, we trained them  for $182$ epochs with a default batch size of $128$ and an initial default learning rate of $1$ which decayed by a factor of $10$ at epochs $91$ and $136$. Optimization was done with stochastic gradient descent using $0.9$ momentum. We used batch normalization, $0.0002$ weight decay, and data augmentation with random cropping and flipping (except for our variants trained on randomly labeled data). We generate results using the 10,000 image testing set.

\noindent \textbf{MLPs (synthetic uncorrelated data):} We use simple multilayer perceptrons with either $10$ or $10,000$ dimensional inputs and binary output. They contained a single hidden layer with $128$ units for the 1x model size and a bias unit. All hidden units were ReLU activated. Weights were initialized using a Gaussian distribution with default standard deviation of $0.01$. Each was trained by default using stochastic gradient descent with momentum of $0.9$ for $50$ epochs on $1,000$ examples produced by a 1/4x sized teacher network with the same architecture which was verified to produce each output for between $40\%$ and $60\%$ of random inputs. Results are generated using a 1,000 image testing set. 

\subsection{Number of Parameters}

\begin{figure*}[]
\centering
\begin{tabular}{@{\hspace{-0.2cm}}c@{\hspace{-0.1cm}}c@{\hspace{-0.1cm}}c}
	\includegraphics[width=.33\linewidth]{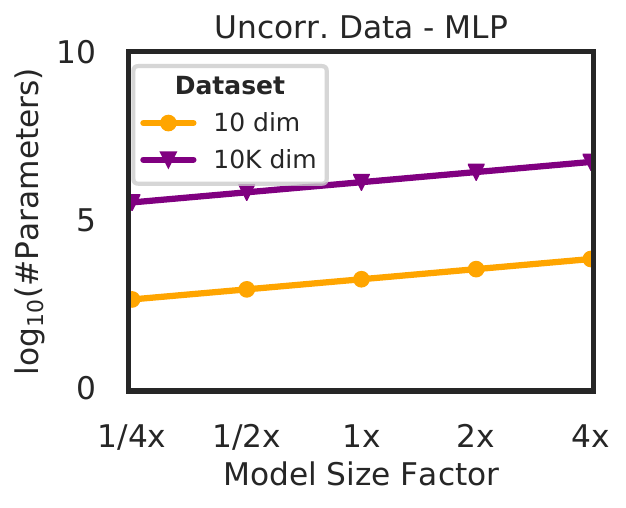}&
	\includegraphics[width=.33\linewidth]{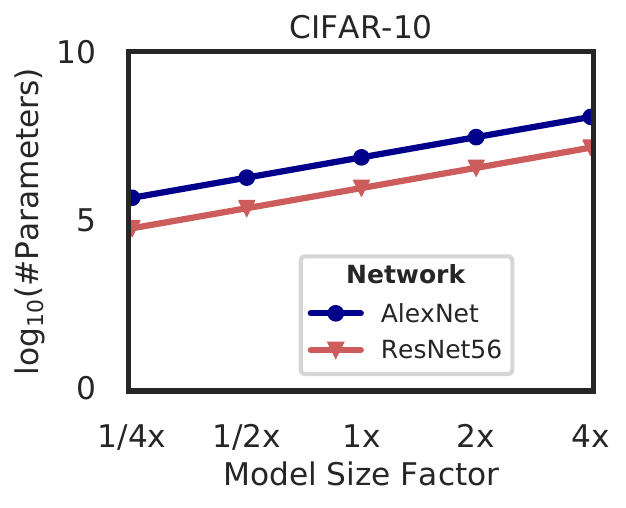} &
	\includegraphics[width=.33\linewidth]{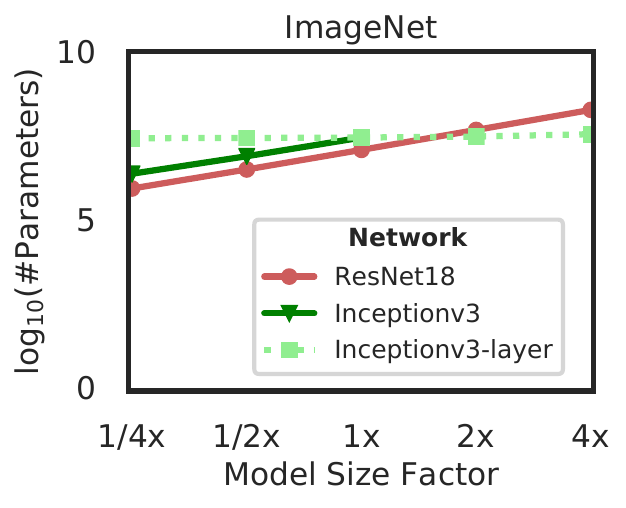}\\
	(a) & (b) & (c) \\
\end{tabular}
\caption{Parameters: (a) Multilayer perceptrons, (b) AlexNets and ResNet56s, (c) ResNet18s, Inception-v3s, and Inception-v3s with a single layer varied. The log number of trainable parameters at each model size.}
\label{fig:params}
\end{figure*}

In Fig.~\ref{fig:params}, we show the number of trainable parameters for each network, showing that they increase on exponential order with the model size factor.

\subsection{Samplings and Replicates}

Due to the number of units in the models and the size of the datasets, analyzing all activations for convolutional filters was intractable in experiments involving redundancy. Instead of an exhaustive sampling, we based our measures on a sampling of spatial locations for each filter capped at 50,000 across the testing set. We ran three independent samplings using this method and found that the variance between them is negligible for all networks.

For all non-ImageNet networks, we conduct three trials with independently trained networks, and plot error bars giving the standard deviations. For redundancy in ImageNet, error bars reflect standard deviations between samplings of units. For prunability in ImageNet, points reflect single trials. 

\section{Additional Results (Section 5)}

\subsection{Non-prunable and non-redundant units increase in quantity but with $g < 2$}

\begin{figure*}[]
\centering
\begin{tabular}{@{\hspace{-0.2cm}}c@{\hspace{-0.1cm}}c@{\hspace{-0.1cm}}c@{\hspace{-0.1cm}}c}
  \includegraphics[width=.26\linewidth]{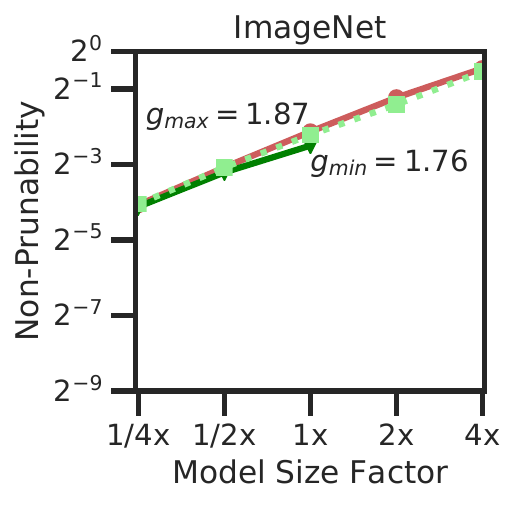} &
  \includegraphics[width=.26\linewidth]{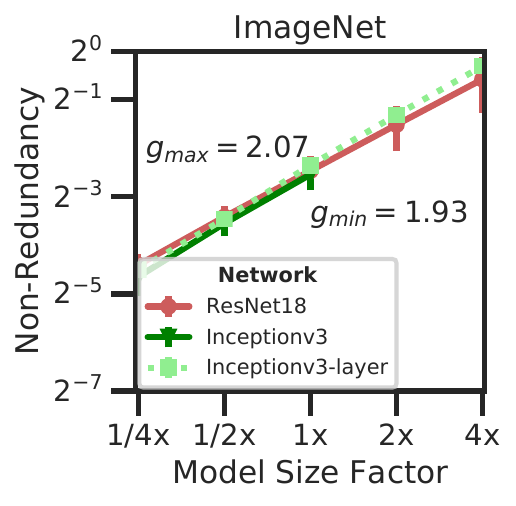} &
    \includegraphics[width=.26\linewidth]{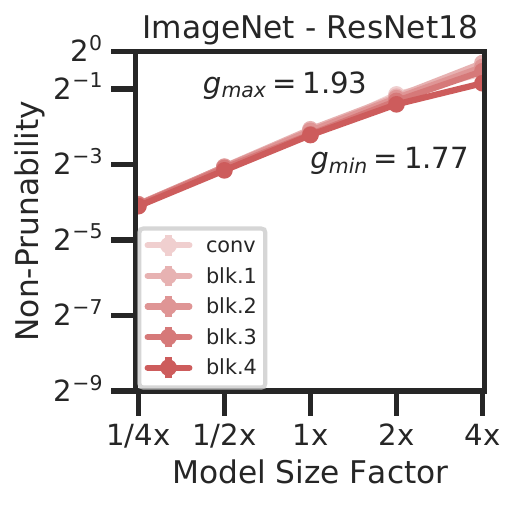} &
    \includegraphics[width=.26\linewidth]{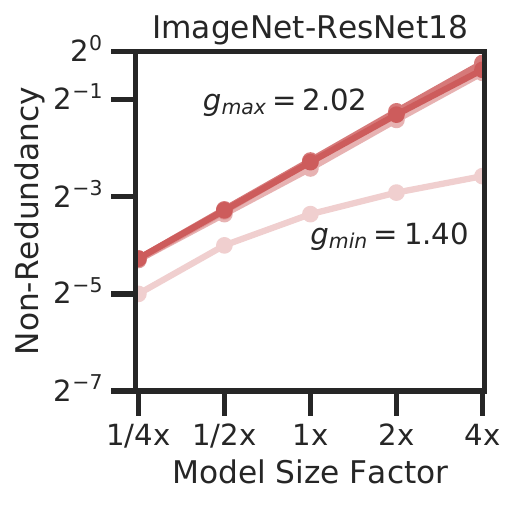}\\
	(a) & (b) & (c) & (d) \\
\end{tabular}
\caption{Non-frivolous units emerge in ResNet18s and Inception-v3s with smaller gains than frivolous ones. (a) Non-prunability. (b) Non-redundancy. (c) Non-prunability layerwise for ResNet18s. (d) Non-redundancy layerwise for ResNet18s. The gain, $g$, gives the increase when the network size is doubled. Max and min $g$ values are given.}
\label{fig:non_imagenet}
\end{figure*}

\begin{figure*}[]
\centering
\begin{tabular}{@{\hspace{-0.2cm}}c@{\hspace{-0.1cm}}c@{\hspace{-0.1cm}}c@{\hspace{-0.1cm}}c}

    \includegraphics[width=.26\linewidth,valign=T]{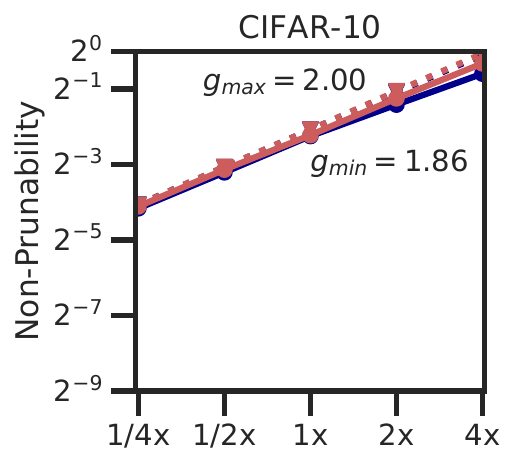} & 
	\includegraphics[width=.26\linewidth,valign=T]{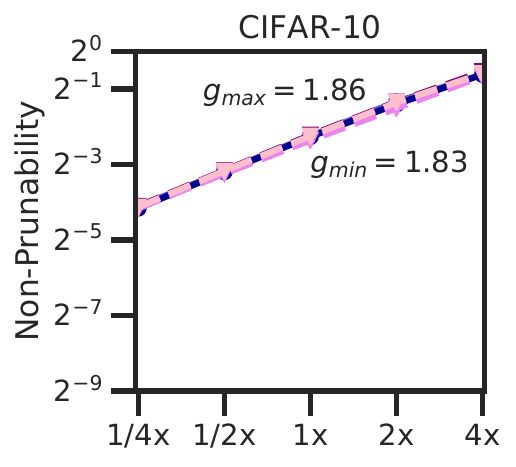}&
	\includegraphics[width=.26\linewidth,valign=T]{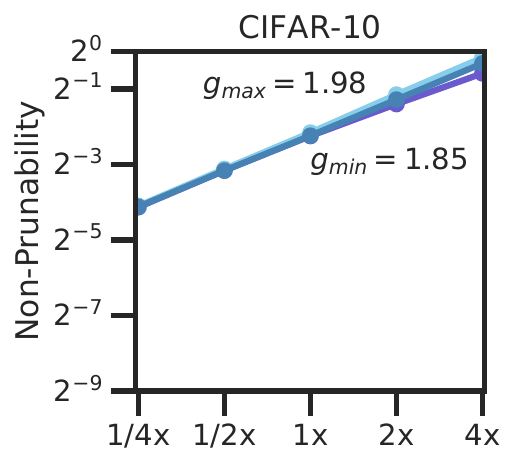} &
	\includegraphics[width=.26\linewidth,valign=T]{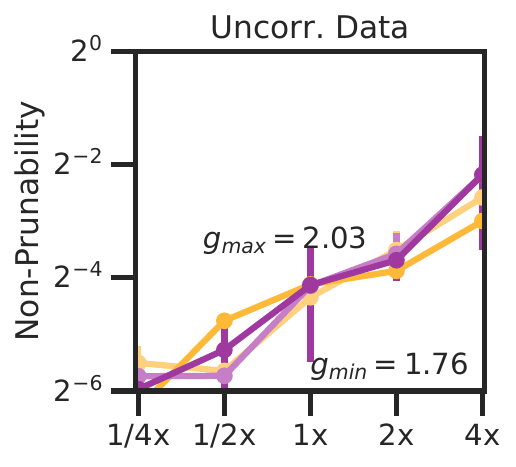}\\

    \includegraphics[width=.26\linewidth]{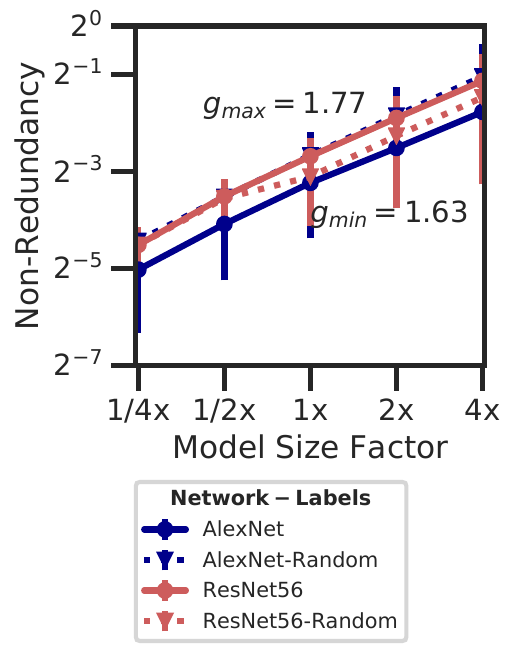} & 
	\includegraphics[width=.26\linewidth]{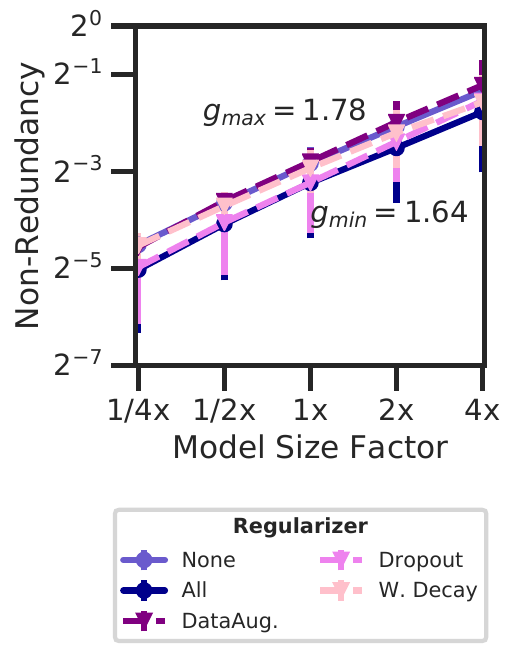}&
	\includegraphics[width=.26\linewidth]{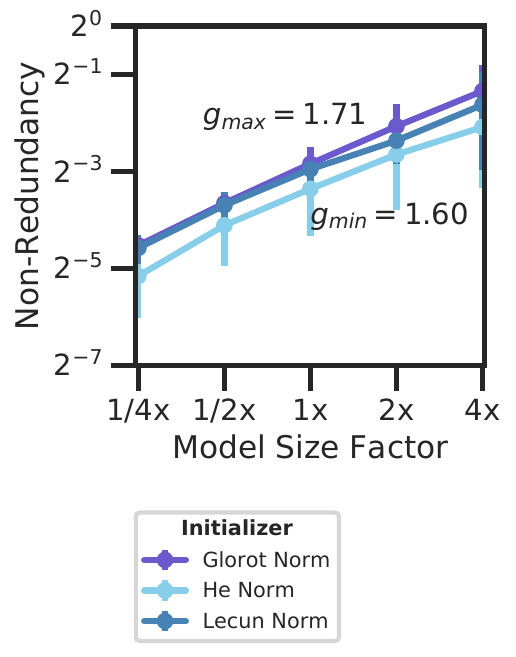} &
	\includegraphics[width=.26\linewidth]{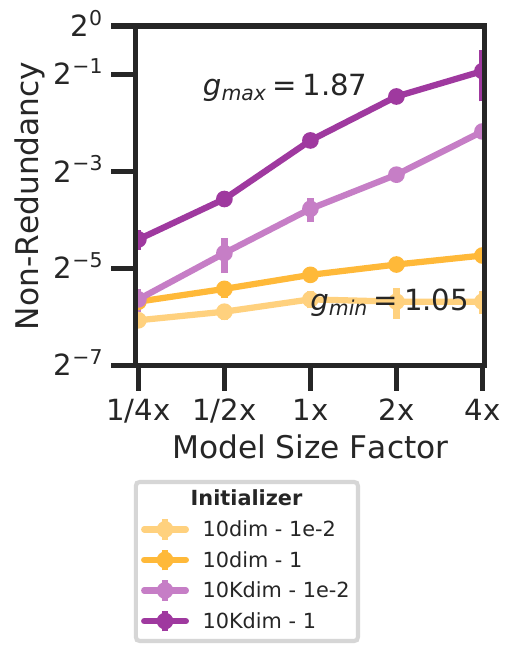}\\
	
	(a) & (b)& (c)& (d)\\
	
\end{tabular}
\caption{Non-frivolous units emerge with smaller gains than frivolous ones for different architectures, regularizers, initializers and datasets. (a) AlexNets and ResNet56s trained with and without random labels (CIFAR-10). (b) AlexNets trained with and without regularization (CIFAR-10). (c) AlexNets trained with various initializations (CIFAR-10). (d) MLPs with different input sizes and initializations (synthetic data). Max and min gain factors are given for each plot.}
\label{fig:non_other}
\end{figure*}

In Fig.~\ref{fig:non_imagenet} and Fig.~\ref{fig:non_other}, we display plots analogous to those presented in the main paper in Fig.~\ref{fig:imagenet_results} and Fig.~\ref{fig:results_other} but plot trends in non-prunable and non-redundant units. As is mirrored by the fact that frivolous units tend to more-than-double, their complements tend to less-than-double when model width doubles. While the gain values are very small for some MLPs, these non-frivolous units increase in quantity for all networks. A compelling direction for future work will be to analyze prunability and redundancy together or alternative types of capacity constraints to see how little the number of non-frivolous components of deep networks can be shown to increase.

\subsection{Layerwise analysis for Inception-v3 models}

\begin{figure*}[]
\centering
\begin{tabular}{c@{\hspace{-0.1cm}}c}
	\includegraphics[width=.3\linewidth]{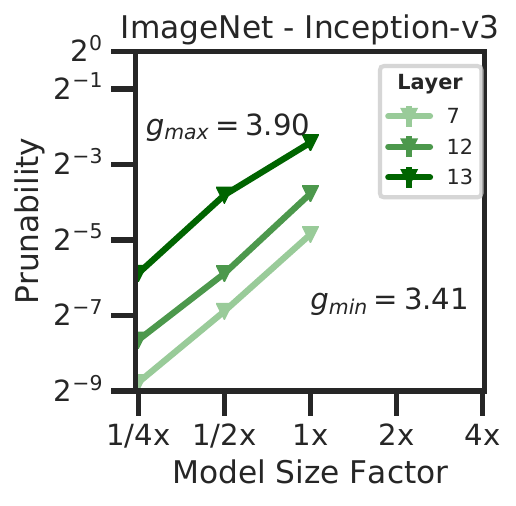} &
	\includegraphics[width=.3\linewidth]{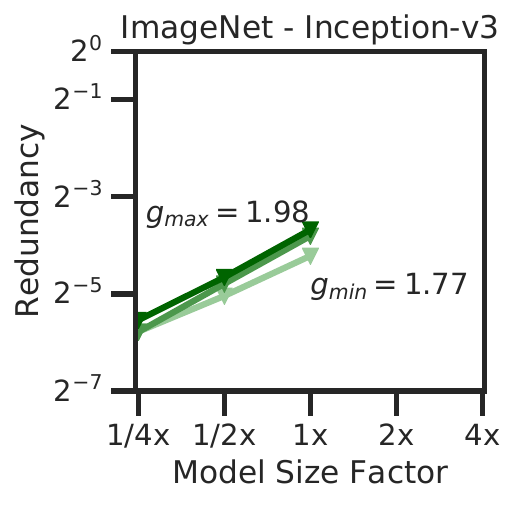} \\
	(a) & (b)\\
\end{tabular} 
\caption{Inception-v3 prunability and redundancy layerwise (ImageNet): Trends in (a) prunability and (b) redundancy among the final $35 \times 35$ (layer 7), $17 \times 17$ (layer 12), and $8 \times 8$ (layer 13) blocks within Inception-v3 in ImageNet. Gains for the individual layers are similar to the network as a whole.}
\label{fig:layerwise_Inception-v3}
\end{figure*}

\begin{figure*}[]
\centering
\begin{tabular}{c@{\hspace{-0.1cm}}c}
	\includegraphics[width=.3\linewidth]{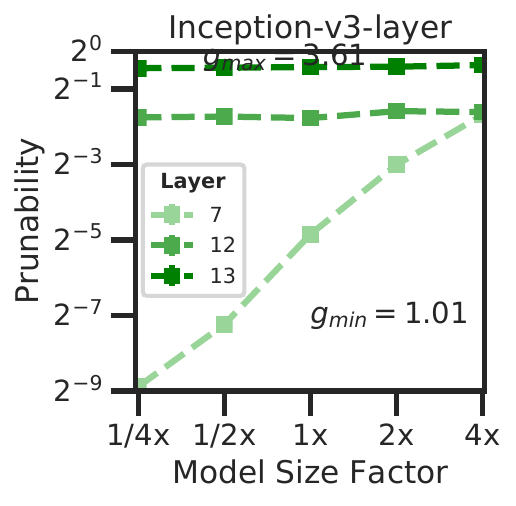} &
	\includegraphics[width=.3\linewidth]{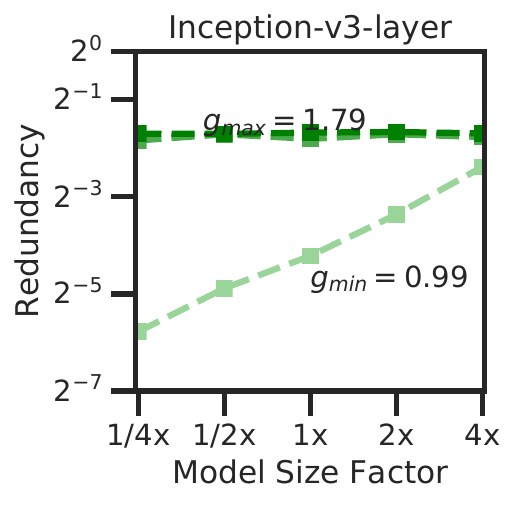} \\
	(a) & (b)\\
\end{tabular}
\caption{Inception-v3 single layer prunability and redundancy layerwise (ImageNet): Trends in (a) prunability, and (b) redundancy among the final $35 \times 35$ (layer 7), $17 \times 17$ (layer 12), and $8 \times 8$ (layer 13) blocks within Inception-v3s in ImageNet as only the final $35 \times 35$ layer is varied in size. Varying the size of a single layer has no effect on the unit prunability and redundancy of other layers.}
\label{fig:layerwise_Inception-v3_layer}
\end{figure*}

Fig.~\ref{fig:layerwise_Inception-v3} and Fig.~\ref{fig:layerwise_Inception-v3_layer} add to the analysis presented in the main paper in Fig.~\ref{fig:imagenet_results}c-d. We show that individual layers in our Inception-v3s trained in ImageNet display unique trends and that even when frivolity increases with $g>2$ for a network as a whole, it does not imply that it does so for all layers. We find that when a single layer is varied keeping others constant, frivolity only increases for that individual layer.

\subsection{Weight initialization and input dimensionality experiments}

\begin{figure*}[]
\centering
\begin{tabular}{ccc}
	\includegraphics[width=.3\linewidth]{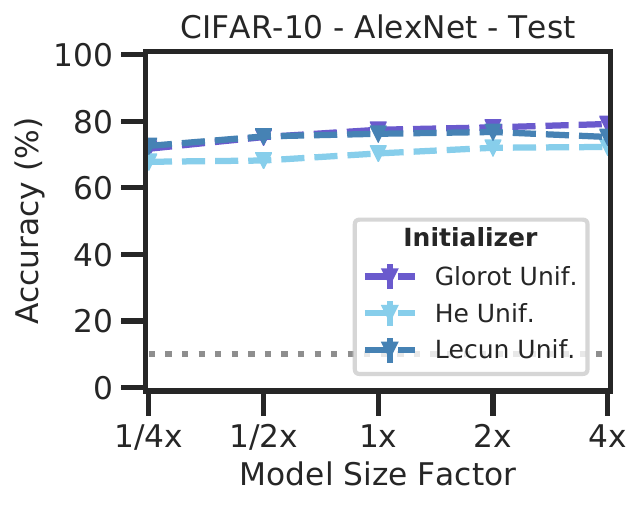}&
	\includegraphics[width=.3\linewidth]{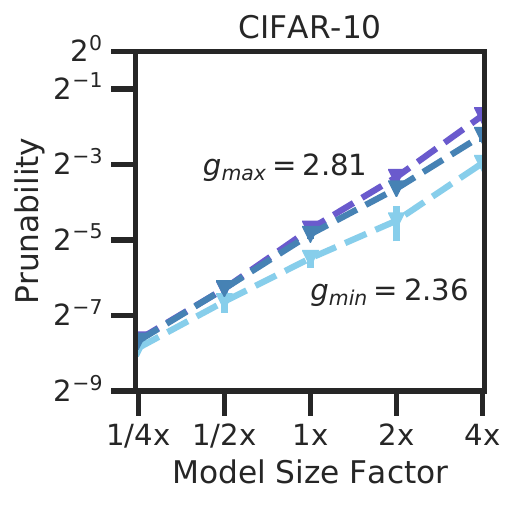} &
	\includegraphics[width=.3\linewidth]{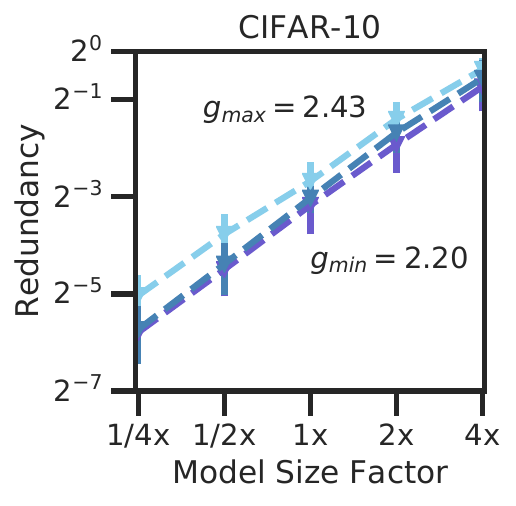} \\
	(a) & (b) & (c)\\
\end{tabular}
\caption{AlexNet prunability and redundancy trends with uniform initializations (CIFAR-10). Trends in (a) accuracy, (b) prunability, and (c) redundancy with Glorot, He, and LeCun uniform initializations across size factors for AlexNets. Trends resemble those for Gaussian initializations.}
\label{fig:init_cifar}
\end{figure*}

In addition to using Gaussian initializations, we also test AlexNets with Uniform Glorot, He, and LeCun initializations to better understand the role of initialization. Glorot initialization~\cite{glorot2010understanding} assigns weights with variance $\sigma^2 = 2/(\textrm{fan\_in}+\textrm{fan\_out})$, LeCun initialization with $\sigma^2=1/\textrm{fan\_in}$ \cite{lecun2012efficient}, and He initialization with $\sigma^2=2/\textrm{fan\_in}$ \cite{he2015delving}. Fig.~\ref{fig:init_cifar} shows that their results are very similar to those of the networks with Gaussian distributed initial weights in Fig.~\ref{fig:results_other}c, suggesting that the initialization distribution matters little compared to the initialization variance.

\begin{figure*}[]
\centering
\begin{tabular}{ccc}
	\includegraphics[width=.3\linewidth]{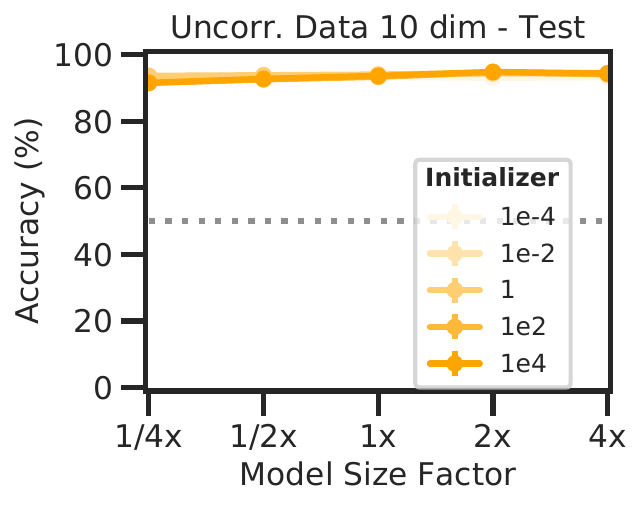}&
	\includegraphics[width=.3\linewidth]{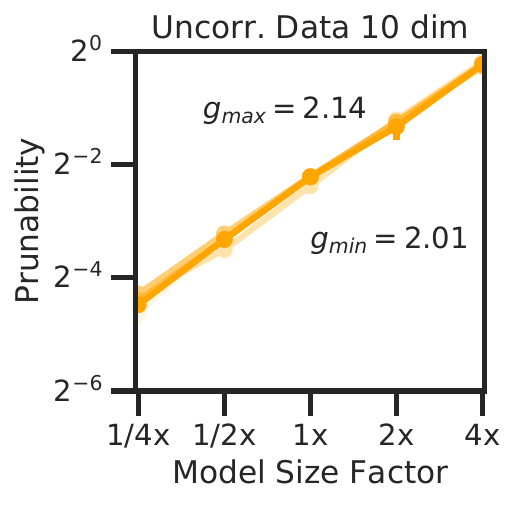} &
	\includegraphics[width=.3\linewidth]{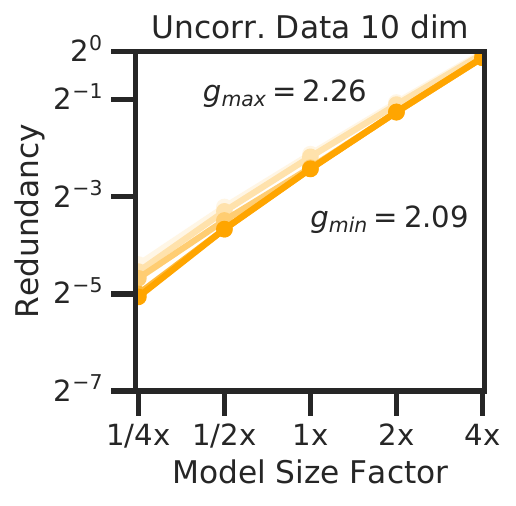} \\
	(a) & (b) & (c)\\
\end{tabular}
\caption{Frivolous units are not sensitive to initialization variance in MLPs trained on 10 dimensional data. Trends in (a) accuracy, (b) prunability, and (c) redundancy with multiple initialization variances across size factors for MLPs trained on synthetic uncorrelated data. The legend gives standard deviations for the Gaussian initialization.}
\label{fig:init_mlp_old}
\end{figure*}

\begin{figure*}[]
\centering
\begin{tabular}{ccc}
	\includegraphics[width=.3\linewidth]{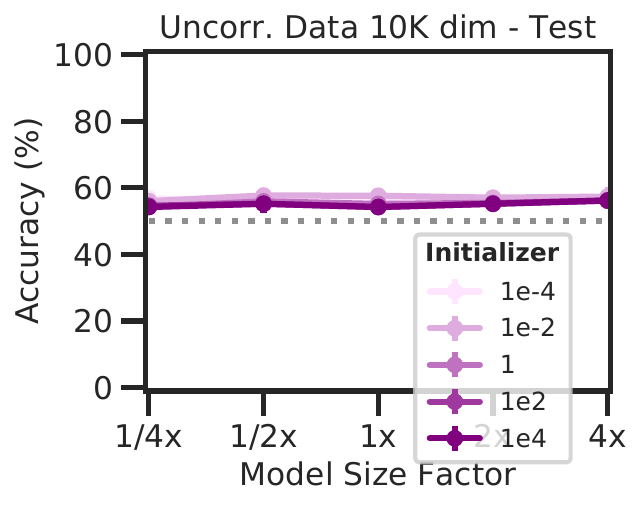}&
	\includegraphics[width=.3\linewidth]{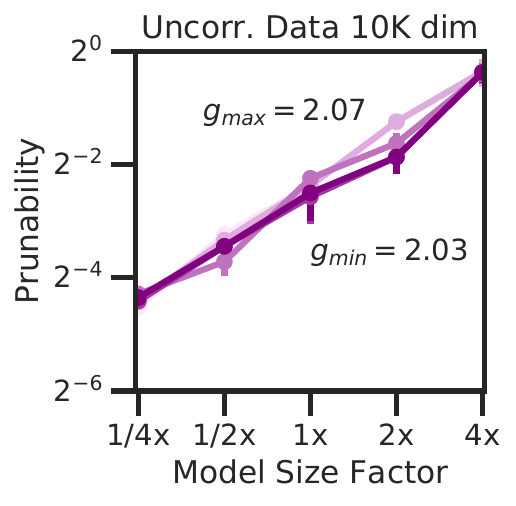} &
	\includegraphics[width=.3\linewidth]{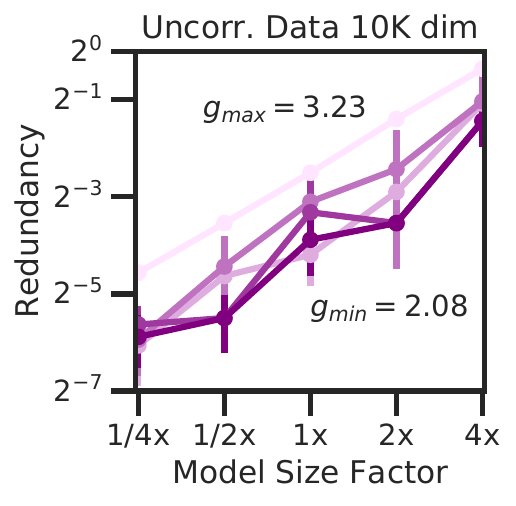} \\
	(a) & (b) & (c)\\
\end{tabular}
\caption{Redundancy is sensitive to initialization variance in MLPs trained on 10,000 dimensional data. Trends in (a) accuracy, (b) prunability, and (c) redundancy with multiple initialization variances across size factors for MLPs trained on synthetic uncorrelated data. The legend gives standard deviations for the Gaussian initialization. Redundancy is sensitive to initialization variance with the smallest variances resulting in the greatest amounts of redundancy.}
\label{fig:init_mlp}
\end{figure*}

In Fig.~\ref{fig:init_mlp} and Fig.~\ref{fig:init_mlp_old}, we show the results of altering variance for Gaussian initializations in the MLPs trained on uncorrelated data. For the datasets with 10-dimensional inputs, results were fairly consistent under alternative initializations. For 10,000 dimensional inputs, however, the amount of redundancy developed was sensitive to initialization and was the highest for the smallest initializations. This demonstrates an interactive effect between data and initialization regarding redundancy.

\subsection{Additional experiments with optimizers, learning rates, batch sizes, and number of training epochs}

\begin{figure*}[]
\centering
\begin{tabular}{ccc}
	\includegraphics[width=.3\linewidth]{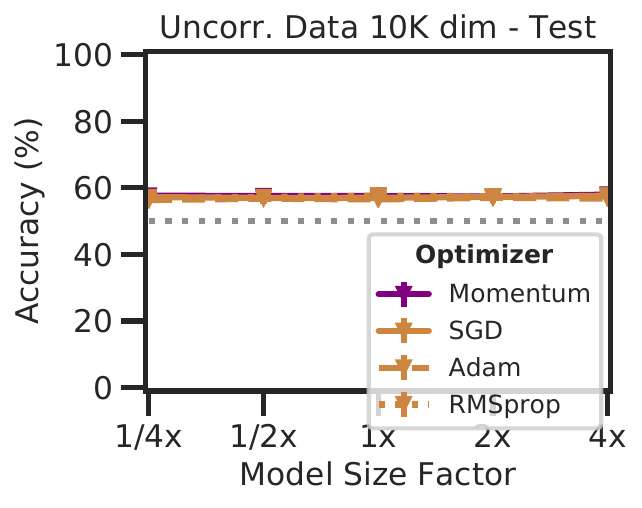}&
	\includegraphics[width=.3\linewidth]{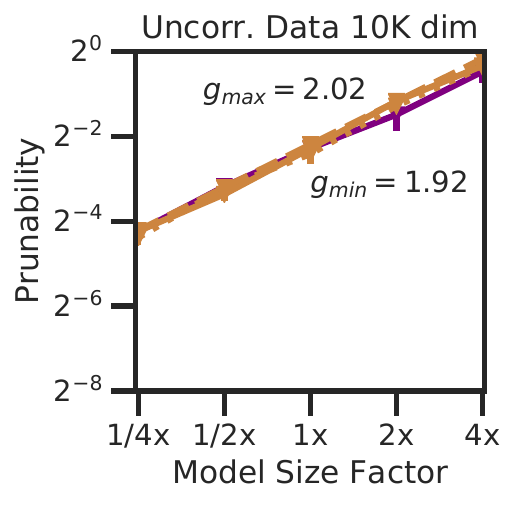}&
	\includegraphics[width=.3\linewidth]{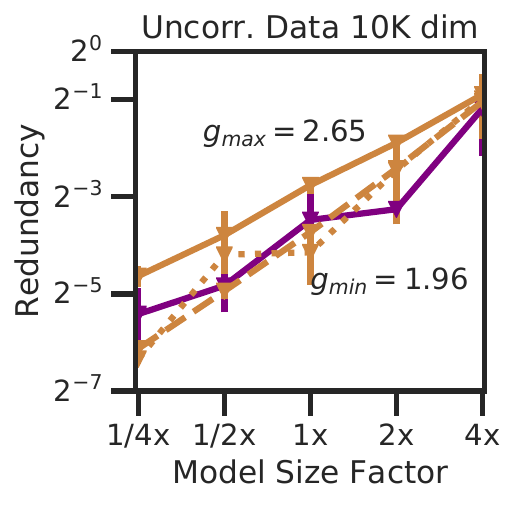} \\
	(a) & (b) & (c) \\
\end{tabular}
\caption{Optimizers influence redundancy in MLPs trained on 10,000 dimensional uncorrelated data. Trends in (a) accuracy, (b) prunability, and (c) redundancy with momentum, stochastic gradient descent, and Adam optimizers across size factors. Momentum, marked in purple, was used for all other experiments with these MLPs. Notably, momentumless stochastic gradient descent resulted in the most redundancy.}
\label{fig:optim_mlp}
\end{figure*}

\noindent {\bf Optimizers.} We test the effect of different optimizers for the MLPs fitting $10,000$ dimensional datapoints. Fig.~\ref{fig:optim_mlp} shows that while prunability is fairly stable under alternate choices of the optimizer, redundancy develops to varying degrees with momentumless stochastic gradient descent resulting in the most.

\begin{figure*}[h!]
\centering
\begin{tabular}{ccc}
	\includegraphics[width=.3\linewidth]{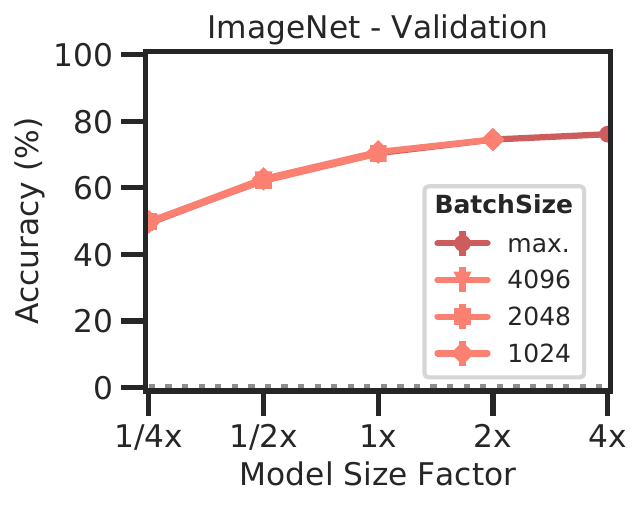}&
	\includegraphics[width=.3\linewidth]{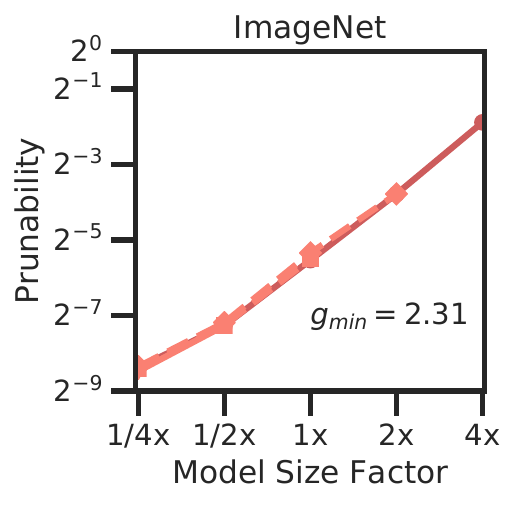}&
	\includegraphics[width=.3\linewidth]{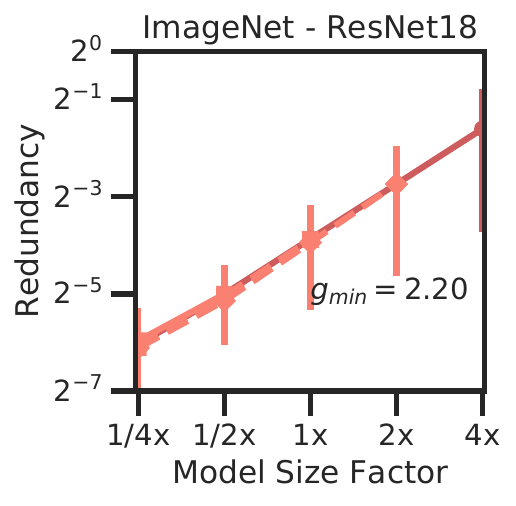} \\
	(a) & (b) & (c)\\
\end{tabular}
\caption{Trends in accuracy and frivolous units do not depend on learning rate and batch size factor in ResNet18s trained in ImageNet. Trends in (a) accuracy, (b) prunability, and (c) redundancy across model sizes. We vary a constant factor $k$ from $1/4$ to $4$ as a multiplier for the batch sizes and learning rates. ``Max'' refers to the maximum batch size that could be used for training a model given available hardware (dgx1 with 8x NVIDIA V100 GPUs 32GB).}
\label{fig:batch_imagenet}
\end{figure*}

\begin{figure*}[]
\centering
\begin{tabular}{ccc}
	\includegraphics[width=.3\linewidth]{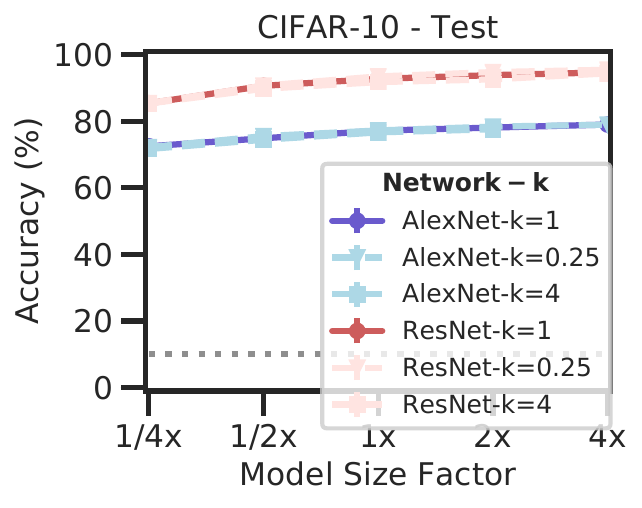}&
	\includegraphics[width=.3\linewidth]{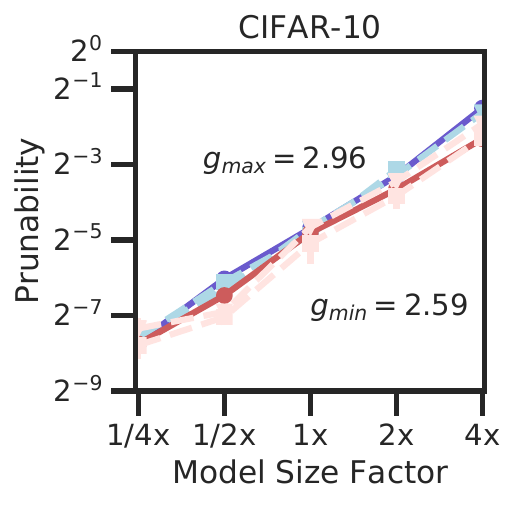}&
	\includegraphics[width=.3\linewidth]{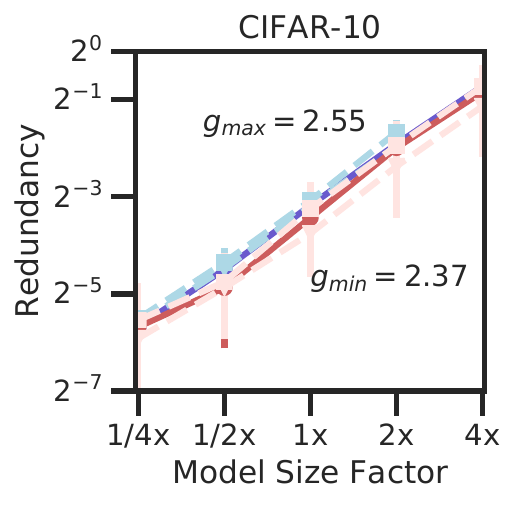}\\
	(a) & (b) & (c)\\
\end{tabular}
\caption{Trends in accuracy and frivolous units do not depend on learning rate and batch size in networks trained on CIFAR-10. Trends in (a) accuracy, (b) prunability, and (c) redundancy across model sizes for AlexNets and ResNet56s. We vary a constant factor $k$ from $1/4$ to $4$ as a multiplier for the batch sizes and learning rates. 4x AlexNets with $k=4$ were not trained due to hardware restrictions. }
\label{fig:batch_cifar}
\end{figure*}

\noindent {\bf Batch Size and Learning Rate.} Batch size and learning rate are commonly varied in practice when training DNNs. To investigate its effects, we vary them by a constant factor, which we denote as $k$. In Fig.~\ref{fig:batch_imagenet}, we report the results for ResNet18 ImageNet and in Fig.~\ref{fig:batch_cifar} for ResNet56 and AlexNet in CIFAR-10. In all tested cases, the batch size and learning rate had no discernible effects on the trends in prunability and redundancy.

\begin{figure*}[]
\centering
\begin{tabular}{ccc}
	\includegraphics[width=.3\linewidth]{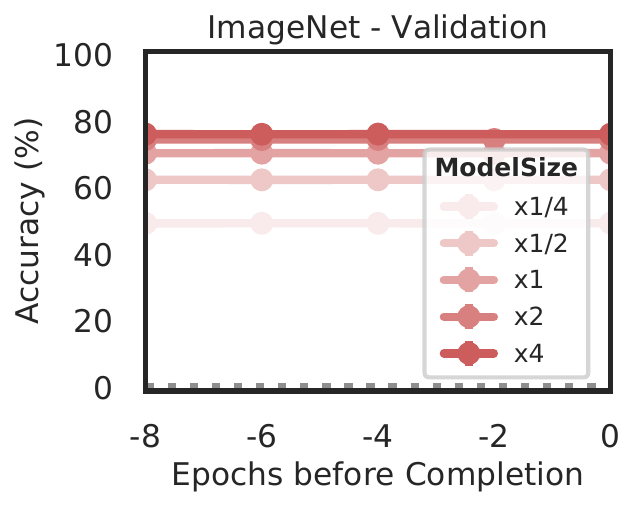}&
	\includegraphics[width=.3\linewidth]{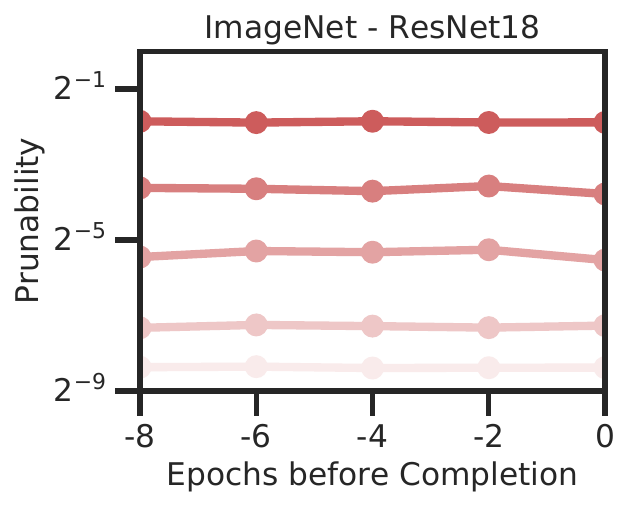} &
	\includegraphics[width=.3\linewidth]{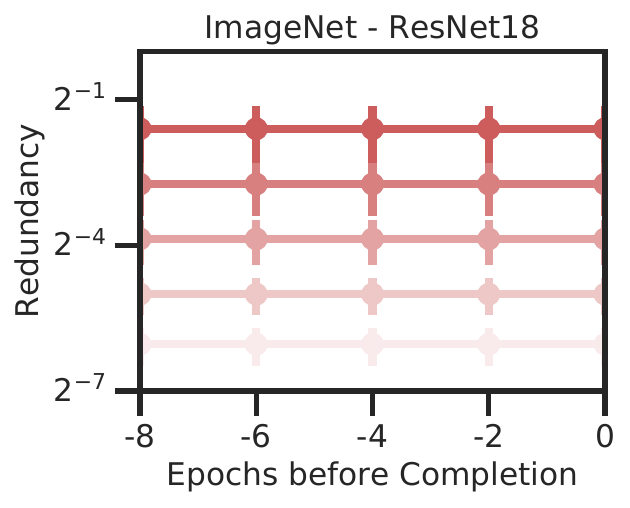}\\
	(a) & (b) & (c)\\
\end{tabular}
\caption{Prunability and redundancy are stable in ResNet18s under different numbers of training epochs past convergence. Trends in (a) accuracy, (b) prunability, and (c) redundancy across training epochs after convergence.}
\label{fig:dynamics}
\end{figure*}

\noindent {\bf Number of Training Epochs.} We find that throughout experiments, trends were robust to changes in the number of training epochs. In Fig.~\ref{fig:dynamics}, we show that trends for prunable and redundant units are invariant to the amount of training time after convergence in ResNet18s.

\subsection{Interpretability Experiments with Lucid}

As described in the main paper, we visualize units in 1/4x and 4x ResNet18s and 1/4x and 1x Inception-v3s to test the hypothesis that the visualizations of units in the thinner networks would be more interpretable due to fewer frivolous units. 

\begin{figure*}[]
\centering
\begin{tabular}{cc}
    (a) & \includegraphics[width=.92\linewidth]{figs/resnet18_block1_thin_lucid.png} \\
     & \includegraphics[width=.92\linewidth]{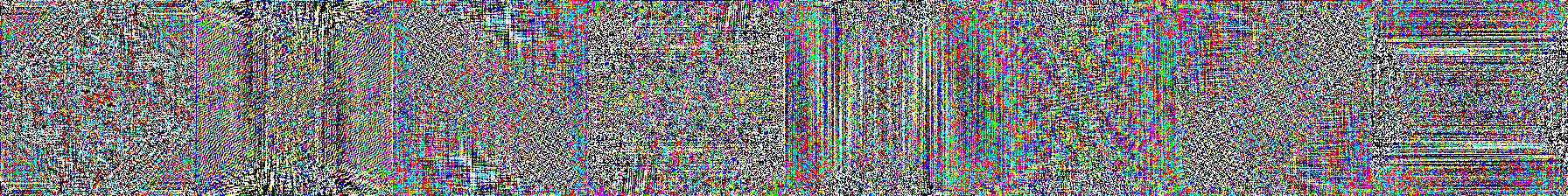} \\
     (b) & \includegraphics[width=.92\linewidth]{figs/resnet18_block1_wide_lucid.png} \\
     & \includegraphics[width=.92\linewidth]{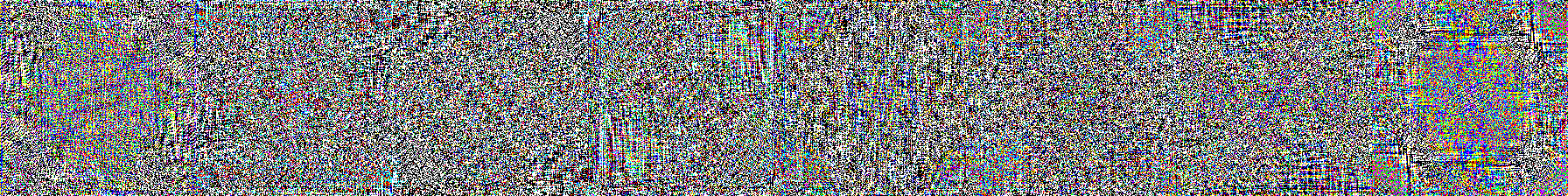}\\
    (c) & \includegraphics[width=.92\linewidth]{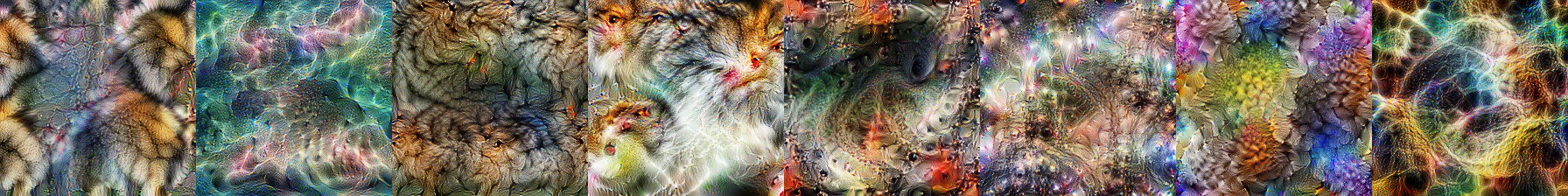} \\
     & \includegraphics[width=.92\linewidth]{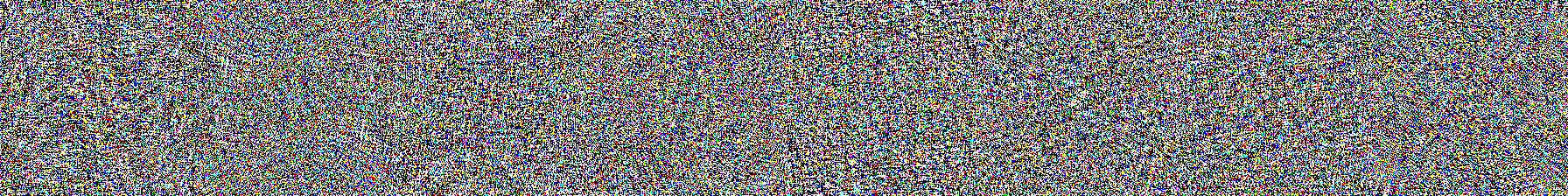} \\
     (d) & \includegraphics[width=.92\linewidth]{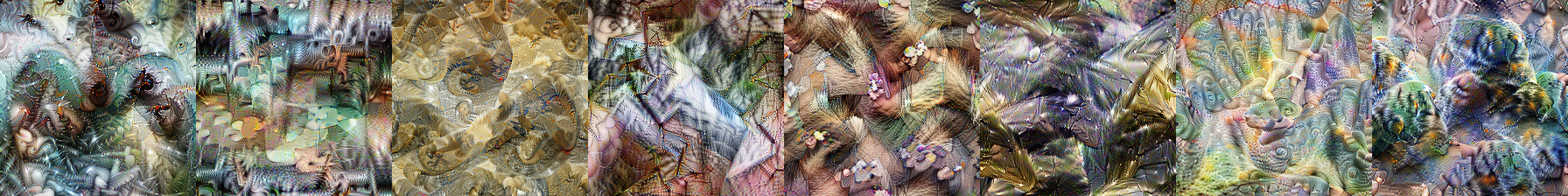} \\
     & \includegraphics[width=.92\linewidth]{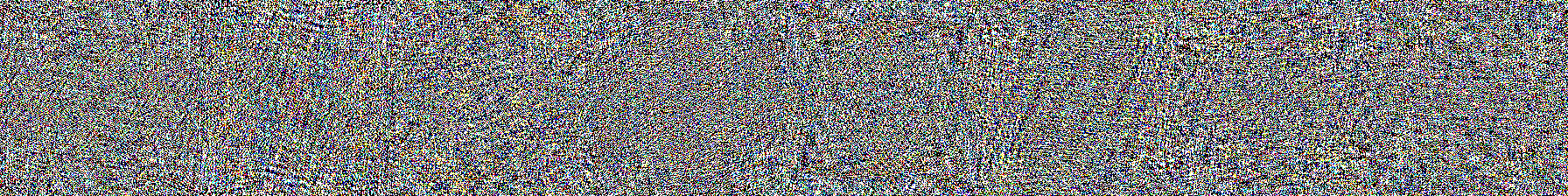}
\end{tabular}
\caption{Visualizations in pixel and Fourier space of 8 units from each of the first block (a,b) and final block (c,d) of the 1/4x (a,c) and 4x (b,d) ResNet18s. The total variations of these images were used for hypothesis testing via a rank-based permutation test..} 
\label{fig:lucid_vis}
\end{figure*}

The images were created using Lucid~\cite{olah2017feature} and were made to optimize images in order to maximize the post-ReLU activation of units from a randomly-chosen set of $8$ at the end of each ResNet18 block and Inception-v3 module. Using Lucid, we parameterized each image in decorrelated Fourier space, and used random padding, jittering, scaling, and rotation during optimization over 256 steps. Example images and their Fourier transforms from the first and final blocks of 1/4x and 4x ResNet18s are shown in Fig.\ref{fig:lucid_vis}.

For the images (both in pixel space and frequency space), our measure of interpretability was an image's total variation which measures the total amount of difference between adjacent pixels across the image. We calculated it by adding together the sum of the absolute valued vertical and horizontal gradients for each channel of the image:
\begin{equation}
\textrm{tvar}(V) = \sum_{c=1}^{3}\left[\; \left|\sum_{ij} \nabla_{vert}(V_c) \right| + \left|\sum_{ij} \nabla_{horiz}(V_c) \right|\; \right],
\end{equation}
where $|\cdot |$ represents the absolute value and the outer sum is over the RGB channels.

When comparing the filter visualizations either in pixel or frequency space from homologous layers in the thin and wide networks, we used a one-sided rank-based permutation test. We calculated the ranks for the total variations of the visualization images from the thin network with respect to the images from both the thin and wide networks and used their sum as the test statistic. Over $100,000$ random shufflings of the ranks, we reported the one-sided $p$ value as the proportion of samplings whose sum was less than or equal to this test statistic. See Table~\ref{tab:vis_pvalues} for these and effect sizes. Note that a small $p$ value indicates that the visualizations of units in the thinner network tend to have a lower total variation (\ie they were more interpretable). While these results show more evidence for the hypothesis that thinner networks are more interpretable than against it, results vary across layers, and there are no consistent trends. 


\begin{table*}[h!]
    \centering
    \footnotesize
    \setlength{\tabcolsep}{2pt} 
    
    \begin{tabular}{|l|r|r|r|r|}
    \hline
        \cellcolor{Gray} \textbf{ResNet18} & \cellcolor{Gray} Block 1 & \cellcolor{Gray} Block 2 & \cellcolor{Gray} Block 3 & \cellcolor{Gray} Block 4 \\ \hline
        \cellcolor{Gray} Total Var p (Pixel) & 0.751 & 0.835 & 0.563 & \textbf{0.001} \\ 
        \cellcolor{Gray} Mean Effect (Pixel) & 1.027 & 1.171 & 0.987 & 0.883 \\ \hline
        \cellcolor{Gray} Total Var p (Freq) & 0.361 & 0.750 & 0.25 & \textbf{0.001} \\ 
        \cellcolor{Gray} Mean Effect (Freq) & 0.970 & 1.127 & 0.967 & 0.863 \\ \hline
    \end{tabular}\\
    
    \vspace{0.1cm}
    
    \begin{tabular}{|l|r|r|r|r|r|r|}
    \hline
        \cellcolor{Gray} \textbf{Inception-v3} & \cellcolor{Gray} Conv4 & \cellcolor{Gray} 35x35x256a & \cellcolor{Gray} 35x35x288a & \cellcolor{Gray} 35x35x288b & \cellcolor{Gray} 17x17x768a & \cellcolor{Gray} 17x17x768b\\ \hline
        \cellcolor{Gray} Total Var p (Pixel) & \textbf{0.982} & 0.566 & 0.080 & \textbf{0.018} & \textbf{0.990} & 0.140 \\ 
        \cellcolor{Gray} Mean Effect (Pixel) & 1.237 & 1.016 & 0.820 & 0.856 & 1.275 & 0.883 \\ \hline
        \cellcolor{Gray} Total Var p (Freq) & \textbf{0.989} & 0.263 & \textbf{0.019} & 0.065 & \textbf{0.998} & 0.118 \\
        \cellcolor{Gray} Mean Effect (Freq) & 1.406 & 0.983 & 0.831 & 0.875 & 1.324 & 0.920 \\ \hline
        
        \cellcolor{Gray} & \cellcolor{Gray} 17x17x768c & \cellcolor{Gray} 17x17x768d & \cellcolor{Gray} 17x17x768e & \cellcolor{Gray} 17x17x1280a & \cellcolor{Gray} 8x8x2048a & \cellcolor{Gray} 8x8x2048b \\ \hline
        \cellcolor{Gray} Total Var p (Pixel) & 0.097 & 0.116 & 0.520 & \textbf{0.010} & \textbf{0.005} & \textbf{0.004} \\
        \cellcolor{Gray} Mean Effect (Pixel) & 0.900 & 0.898 & 0.977 & 0.858 & 0.884 & 0.814 \\ \hline
        \cellcolor{Gray} Total Var p (Freq) & 0.194 & 0.096 & 0.714 & 0.443 & 0.288 & 0.138 \\
        \cellcolor{Gray} Mean Effect (Freq) & 0.943 & 0.926 & 1.018 & 0.943 & 0.973 & 0.917 \\ \hline
    \end{tabular}
    \caption{One-sided p values and mean effect sizes testing for low total variation among visualizations of thin versus wide networks. Each p value is from rank-based permutation test on 8 units from each network and 100,000 samples. Results are given both for images in pixel and frequency space. A small p value indicates that the visualizations of units in the thinner network have a lower total variation. All p values below 0.025 or above 0.975 are bolded. Each effect size gives the mean total variation for the thin networks divided by the same for the wide networks. A small mean effect means that the thin network's visualizations had a smaller average total variation.}
    \label{tab:vis_pvalues}
\end{table*}

\end{document}